\documentclass[10pt,twocolumn,letterpaper]{article}

\usepackage[pagenumbers]{cvpr} % To force page numbers, e.g. for an arXiv version
\usepackage{multirow}
\usepackage{booktabs}
\usepackage{makecell}

\definecolor{cvprblue}{rgb}{0.21,0.49,0.74}
\usepackage[pagebackref,breaklinks,colorlinks,allcolors=cvprblue]{hyperref}

\def\paperID{***} % *** Enter the Paper ID here
\def\confName{CVPR}
\def\confYear{20**}

\title{VABench: A Comprehensive Benchmark for Audio-Video Generation}

\author{%
  \textbf{Daili Hua{$^{1}$}\thanks{These authors contributed equally.}, \ Xizhi Wang{$^6$}\footnotemark[1], \ Bohan Zeng{$^1$}\footnotemark[1], \ Xinyi Huang{$^{1}$}, \ Hao Liang{$^{1,2}$}, \ Junbo Niu{$^{1}$}} \\ \textbf{Xinlong Chen{$^{5}$}, Quanqing Xu{$^{4}$}, \ Wentao Zhang{$^{1,2,3}$}\thanks{Corresponding Author: wentao.zhang@pku.edu.cn}} \\
{\small $^1$Peking University \quad $^2$Zhongguancun Academy \quad $^3$Beijing Key Laboratory of Data Intelligence (Peking University)} \\
    {\small $^4$Ant Group \quad $^5$Institute of Automation, Chinese Academy of Sciences \quad $^6$Huazhong University of Science and Technology} \\ 
}

\begin{document}
\maketitle

{
\def\thefootnote{}\footnotetext{Project Repository: \url{https://github.com/tanABCC/VABench}}
}
\begin{abstract}

Recent advances in video generation have been remarkable, enabling models to produce visually compelling videos with synchronized audio. While existing video generation benchmarks provide comprehensive metrics for visual quality, they lack convincing evaluations for audio-video generation, especially for models aiming to generate synchronized audio-video outputs. To address this gap, we introduce VABench, a comprehensive and multi-dimensional benchmark framework designed to systematically evaluate the capabilities of synchronous audio-video generation. VABench encompasses three primary task types: text-to-audio-video (T2AV), image-to-audio-video (I2AV), and stereo audio-video generation. It further establishes two major evaluation modules covering 15 dimensions. These dimensions specifically assess pairwise similarities (text-video, text-audio, video-audio), audio-video synchronization, lip-speech consistency, and carefully curated audio and video question-answering (QA) pairs, among others. Furthermore, VABench covers seven major content categories: animals, human sounds, music, environmental sounds, synchronous physical sounds, complex scenes, and virtual worlds. We provide a systematic analysis and visualization of the evaluation results, aiming to establish a new standard for assessing video generation models with synchronous audio capabilities and to promote the comprehensive advancement of the field.

\end{abstract}

\section{Introduction}
\label{sec:intro}

\begin{figure*}[t]
    \centering
    \includegraphics[width=0.9\textwidth]{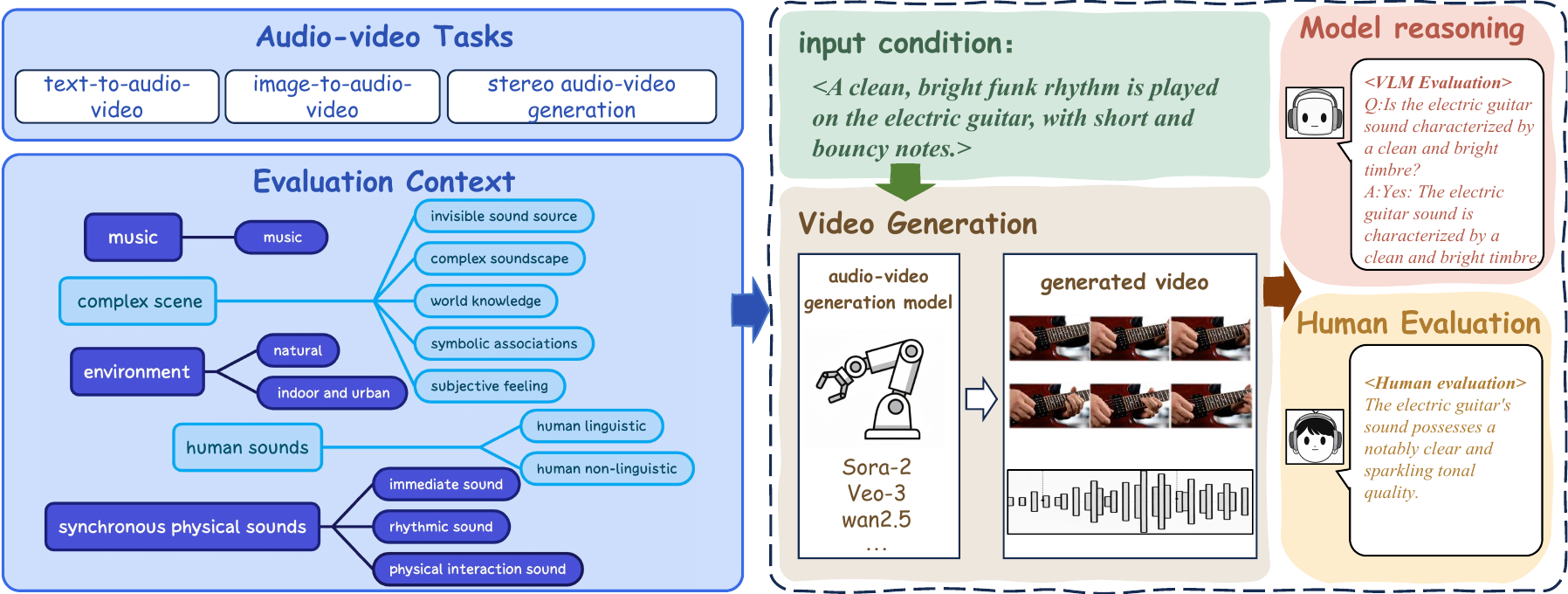}
    \caption{Overview of the VABench framework, illustrating its three main components: (1) The audio-video generation tasks being evaluated (T2AV, I2AV, and stereo), (2) the detailed taxonomy of evaluation contexts (e.g., human sounds, complex scenes), and (3) the evaluation pipeline.}
    \label{fig:teaser}
    % \vspace{-4mm}
\end{figure*}
Video generation technology is rapidly evolving from early content synthesis to intelligent creation, driving innovations in automation across fields such as film production~\cite{polyak2024movie,wu2025automated} and artistic creation~\cite{liu2024video,chen2023control,wu2023tune,guo2023animatediff,an2025unictokens}. Significant progress has been made in pure visual generation regarding resolution and spatiotemporal consistency, supported by well-established evaluation systems (e.g., VBench~\cite{huang2024vbench}, VBench 2.0~\cite{zheng2025vbench}, Evaluation agent~\cite{zhang2025evaluation}). With growing user demands and the inherent coexistence of audio and video in the real world, video generation models with synchronous audio (e.g., Veo~3~\cite{google2025veo3}, Sora~2~\cite{openai2025sora2}, Wan~2.5~\cite{alibaba2025wan25}) are becoming a new technical focus, marking the advent of a truly unified audio-visual generation era.

Despite preliminary achievements, a systematic joint audio-video benchmark remains lacking. Existing works(e.g., JAVISDiT~\cite{liu2025javisdit}), while exploratory, generally suffer from limited evaluation dimensions and constrained scenarios. More critically, they largely overlook unique multimodal coupling phenomena inherent in joint audio-video generation, such as the Doppler effect caused by motion, the synergistic expression of character emotions across audio-visual modalities, and the coordination between background music and visual rhythm. Furthermore, while most current synchronous audio-video models output stereo audio, existing benchmarks lack evaluation tailored to the spatial acoustic properties of such audio. Therefore, there is an urgent need for a comprehensive evaluation system that can balance generation quality, cross-modal semantic consistency, physical plausibility, and emotional expressiveness to address the complex challenges emerging in joint audio-video generation.

To bridge this gap, we introduce VABench, a comprehensive benchmark specifically designed for the holistic evaluation of synchronous audio-video generation. As shown in Fig.~\ref{fig:teaser}, VABench encompasses two primary audio-video generation tasks, text-to-audio-video (T2AV) and image-to-audio-video (I2AV), and features a diverse set of test content covering seven core sound categories including animals, human sounds, music, environmental sounds, synchronous physical sounds, complex scenes, and virtual worlds. This requires models not only to achieve synchronized audio but also to maintain fidelity to real-world internal logic. We employed human workers and large language models to filter testing samples and adjust the distribution of test data. To effectively evaluate audio-video generation performance, VABench incorporates 15 fine-grained metrics, including 8 based on expert models and 7 based on multimodal large language models (MLLMs). These metrics cover critical dimensions such as audio-visual synchronization, lip-speech consistency, and cross-modal similarity, ensuring precise and domain-aware assessment of multimodal generation quality.

Additionally, VABench introduces dual-channel stereo audio evaluation. Addressing the spatial acoustic properties often overlooked by current benchmarks, we provide dedicated test cases to measure the spatial auditory perception and sound field rendering capabilities of generated content. This enables VABench to more comprehensively evaluate audio-video generation capabilities and provides guidance for future joint audio-video generation technologies advancing toward higher-dimensional realism. Our main contributions are as follows:

\begin{itemize}
\item We propose VABench, a comprehensive benchmark for audio-video generation covering two mainstream tasks (T2AV and I2AV). It introduces a suite of 15 fine-grained metrics designed for systematic and effective evaluation.
\item Our benchmark's test set spans seven major content categories. This comprehensive design pushes evaluation beyond simple perceptual coherence, assessing a model's grasp of complex real-world dynamics, including its understanding of physical logic and its ability to capture nuanced human emotional contexts.
\item We also introduce stereo dual-channel audio evaluation into a video generation benchmark. VABench provides dedicated test cases and metrics for spatial audio rendering, making our benchmark more comprehensive and offering guidance for the future development of audio-video generation technologies.
\end{itemize}

\section{Related Works}
\label{sec:related_works}

\subsection{Video Generation Models}
In recent years, diffusion models~\cite{NEURIPS202149ad23d1,he2023latentvideodiffusionmodels,ho2022video,bar2024lumiere,ma2024latte,wang2025lavie,luo2023decomposed,ho2020denoising,wang2023modelscope} and Transformer architectures~\cite{dingCogView2,zhengCogView3,hong2022cogvideolargescalepretrainingtexttovideo,yang2024cogvideox,ma2024latte,yu2023magvit} have jointly driven remarkable progress in video generation. Building upon these foundations, Sora~\cite{openai2024sora} emerged as the first text-to-video model to deliver visually stunning and coherent long-duration results, showcasing the potential of large-scale training despite early challenges in spatial consistency. Subsequently, robust open-source models like Wan~\cite{wan2025wan} set new benchmarks for text-to-video generation, while Seedance~\cite{gao2025seedance} pioneered reinforcement learning from human feedback (RLHF) to greatly improve motion quality and visual fidelity.

\subsection{Audio-Visual Generation}
Video generation has evolved from visual-only synthesis to unified audio–video generation, exemplified by Veo3~\cite{google2025veo3}, Sora2~\cite{openai2025sora2}, and Wan2.5~\cite{alibaba2025wan25}, which pursue precise temporal synchronization between sound and motion. Some methods achieve audio–video synthesis by integrating video-to-audio modules(V2A), making performance largely depend on V2A quality. Recent V2A models like MMAudio~\cite{cheng2025mmaudio} and Kling-Foley~\cite{wang2025kling} improved semantic controllability and event alignment through joint video–text conditioning, while DeepSound-V1~\cite{liang2025deepsound} and Thinksound~\cite{liu2025thinksound} leveraged MLLM-based reasoning for better video understanding and audio fidelity. Yet, maintaining long-term consistency in complex scenes remains challenging.

\subsection{Evaluation Benchmarks}
Evaluation methodologies~\cite{lin2025perceiveanythingrecognizeexplain,an2026genius,luo2024llm} have evolved alongside generation models. For visual content, traditional single metrics such as Inception Score (IS)~\cite{salimans2016improved} and Fréchet Video Distance (FVD)~\cite{unterthiner2019fvd} have been replaced by multi-dimensional frameworks. With the development of large models~\cite{luo2024llm, lin2024draw, xu2025qwen2, lin2025perceiveanythingrecognizeexplain, an2025unictokens, chen2025versavid, shi2025mavors, bai2025multi}, VBench~\cite{huang2024vbench} and VBench2.0~\cite{zheng2025vbench}, cover frame-level quality, temporal consistency, and higher-level realism like physical plausibility. While important for visual evaluation, they lack mechanisms to assess cross-modal consistency.

Compared with the rapid progress in generation methods, evaluation benchmarks for joint audio-visual synthesis remain underdeveloped. Existing benchmarks~\cite{guo2025brace, shi2025mme}, mostly for V2A tasks, are reference-based, requiring real video and audio ground truth—an approach unsuitable for T2AV generation, which instead needs reference-free evaluation of triangular consistency among text, video, and audio. Current evaluations are also constrained by (1) reliance on manual assessment (e.g., Movie Gen~\cite{polyak2024movie}), limiting scalability, and (2) limited coverage in holistic benchmarks (e.g., JavisBench~\cite{liu2025javisdit}), lacking quantitative measures for higher-order couplings such as physical or emotional coherence. To address these gaps, VABench introduces a fine-grained, automated framework for multi-dimensional evaluation of audio-video generation models.
\section{VABench}
\label{sec:vabench}
In this section, we present the details of VABench. ~\cref{subsec:data_category} details our three tasks and data categories. ~\cref{subsec:data_collect} briefly outlines the data generation and collection methods. ~\cref{subsec:eval_metrics} introduces the five levels of evaluation metrics and their specific implementation details.

\subsection{Data Category}
\label{subsec:data_category}

\subsubsection{Task Category}

\noindent \textbf{Text-to-Audio-Video Generation (T2AV).} Translating textual semantics into coherent audio-visual sequences. Key challenges include achieving high-fidelity motion consistency and precise cross-modal semantic alignment.

\noindent \textbf{Image-to-Audio-Video Generation (I2AV).} Inferring motion and synchronized audio from a static image, with key challenges in ensuring action plausibility, temporal coherence, and precise audio-visual alignment.

\noindent \textbf{Stereophonic Audio Generation.} Converting text into stereo audio with explicit spatial cues, challenged by accurate spatial interpretation and channel separation. Evaluation uses 116 prompts specifying distinct left/right sounds.

\subsubsection{Content Category}
\begin{figure}[t]
    \centering
    \includegraphics[width=0.92\linewidth]{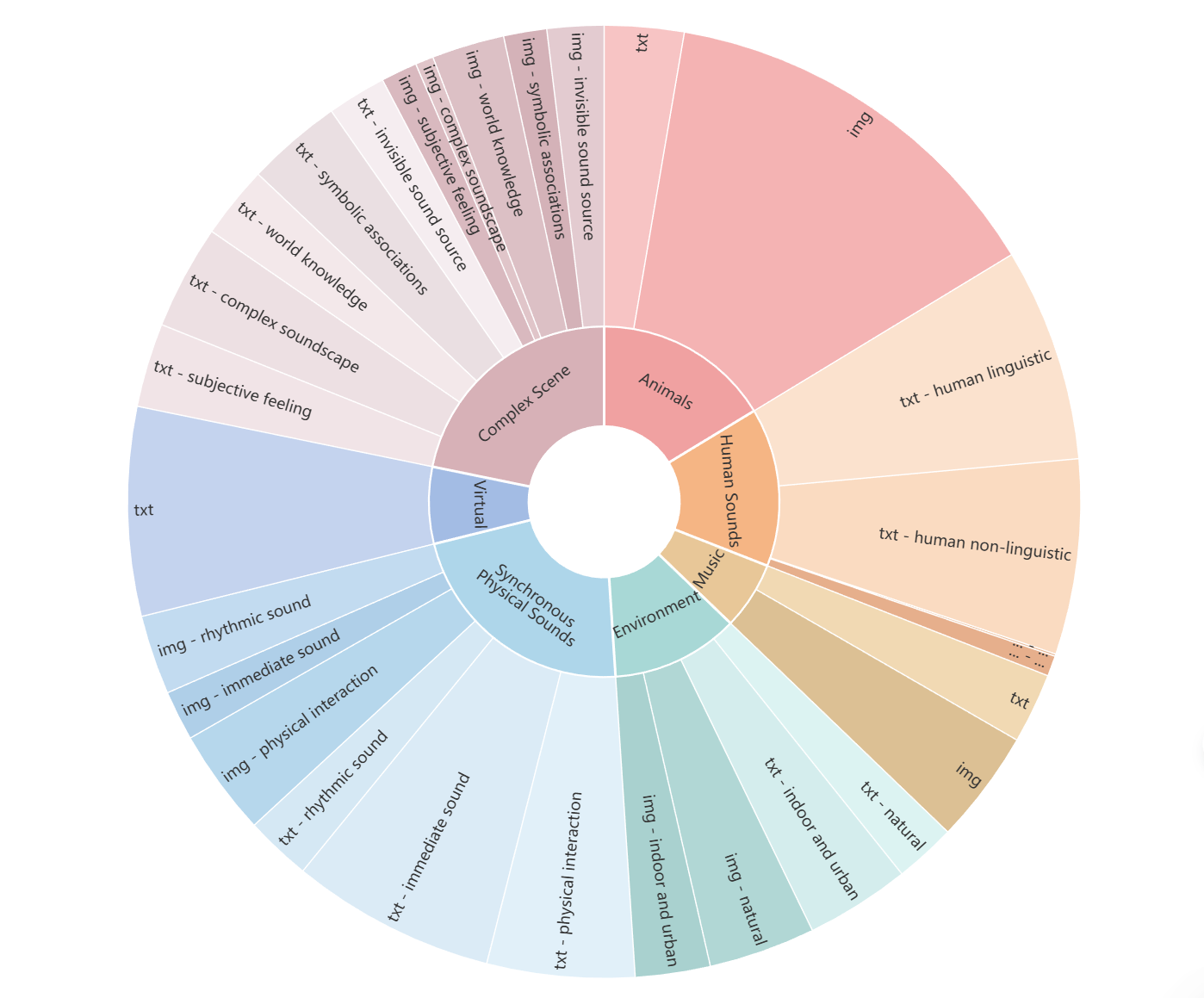}
    \caption{Data distribution of VABench. The sunburst chart illustrates the hierarchical breakdown of our dataset across the seven major content categories and their sub-divisions.}
    \label{fig:data distribution}
\end{figure}

\begin{figure}[t]
    \centering
    \includegraphics[width=0.92\linewidth]{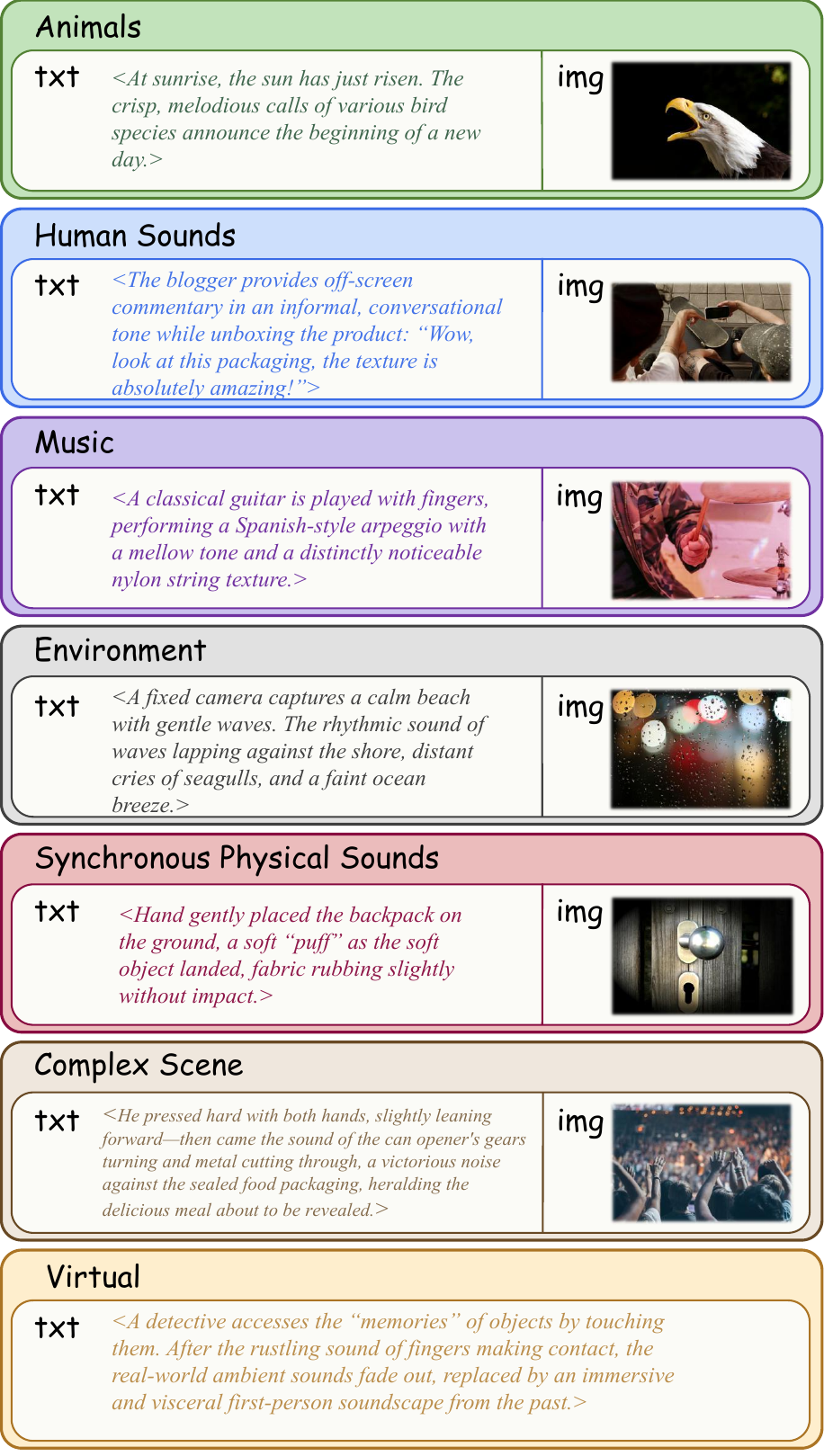}
    \caption{VABench's seven content categories, illustrated with example text prompts and representative images.}
    \label{fig:test_cases}
\end{figure}

To evaluate sound realism, A-V consistency, and semantic plausibility, we developed a seven-category taxonomy grounded in human auditory perception. Its rigor is shown as our dimensions encompass Kling-Foley-Eval's~\cite{wang2025kling} acoustic classes and address physical plausibility—a core principle aligned with VBench~2.0's~\cite{zheng2025vbench} visual-centric focus. The taxonomy spans basic sources, physical interactions, complex semantics, and non-realistic content, organized as follows.

\noindent \textbf{Animals.} This category encompasses vocalizations across diverse species, focusing on accurate species-specific acoustic modeling and audio-visual behavioral consistency.

\noindent \textbf{Human Sounds.} This category is subdivided into linguistic sounds involving semantic content, and non-linguistic sounds related to physiological states or actions.

\noindent \textbf{Music.} This category covers structured audio content across genres. Evaluation focuses on melodic and rhythmic coherence, timbre authenticity, and alignment with visual emotional tone.

\noindent \textbf{Environmental Sounds.} This category integrates three major soundscapes: natural, urban, and indoor environments.

\noindent \textbf{Synchronous Physical Sounds.}
This category targets immediate, rhythmic, or physical interaction sounds, demanding strict adherence to material properties and motion dynamics for precise audio-visual alignment.

\noindent \textbf{Complex Scenes.}
This category targets high-order scenarios spanning five dimensions: complex soundscapes, subjective feelings, world knowledge, symbolic associations, and invisible sound sources, demanding synergistic audio-visual reasoning.

\noindent \textbf{Virtual Worlds.} This category targets non-realistic scenarios  that transcend physical laws yet demand internal logical and stylistic coherence. Due to the unique nature of this category, it is exclusively featured in the T2AV task.

To make our VABench more intuitive, we provide specific test cases in ~\cref{fig:test_cases}.

\subsection{Data Collection}
\label{subsec:data_collect}

\begin{figure}[t]
    \centering
    \includegraphics[width=0.92\linewidth]{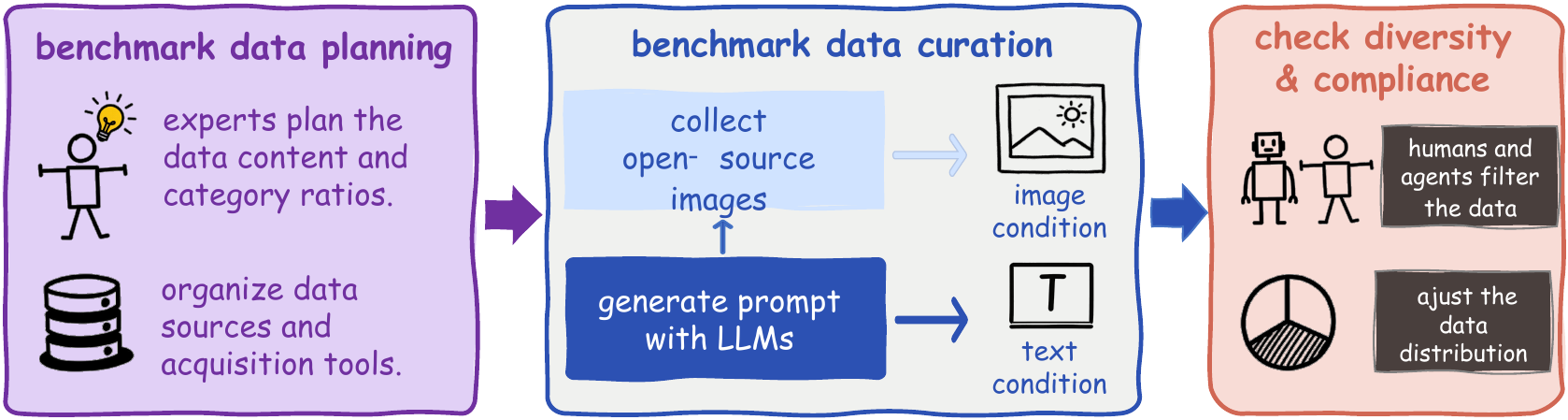}
    \caption{Overview of the pipeline for benchmark data curation. This process is used to generate the text conditions for T2AV tasks and the image conditions for I2AV tasks.}
    \label{fig:data_pipeline}
\end{figure}

We employ a dual-path strategy (T2AV and I2AV) to build a high-fidelity dataset spanning seven categories, comprising 778 T2AV and 521 I2AV samples (\cref{fig:data distribution}). Both pipelines utilize LLMs and VLMs to generate structured prompts and QA pairs, followed by rigorous human verification to ensure semantic accuracy and audio-visual consistency (\cref{fig:data_pipeline}).

\noindent \textbf{Text-to-Audio-Video (T2AV).} We use expert templates and LLMs to batch-generate raw prompts, which are then used to create visual question-answer pairs (VQA) and audio question-answer pairs (AQA) pairs for evaluation. An LLM also structurally decouples these prompts into visual and auditory sub-prompts. A final human verification step ensures correct categorization, element observability, and adherence to physical/commonsense constraints.

\noindent \textbf{Image-to-Audio-Video (I2AV).} We curate and manually classify high-quality images, carefully excluding any content with privacy concerns. An MLLM then generate unified audio-visual descriptions detailing objective visuals and commonsense-inferred audio. These descriptions are simultaneously used to construct VQA/AQA pairs (verifying fidelity and plausibility) and decoupled by an LLM into sub-prompts. Human rechecks then validate auditory inferences and question discriminability.

\subsection{Evaluation Metrics}
\label{subsec:eval_metrics}

To comprehensively assess generated content, we propose a dual-track framework combining specialized precision with holistic understanding. It includes two components: Expert Model-based Evaluation, which quantifies perceptual quality via specialized models, and MLLM-based Evaluation, which simulates human judgment of complex audio-visual semantics.

\subsubsection{Expert Model-based Evaluation} 
To assess high-quality synergy among text, visual, and audio modalities, our framework evaluates three critical dimensions: uni-modal quality, cross-modal semantic alignment, and temporal synchronization.

\noindent \textbf{Uni-modal Audio Quality - SpeechClarity.} 
Assesses background noise and perceptual speech quality for the human linguistic sub-category. For implementation, we leverage the Overall Quality (OVRL) score from DNSMOS~\cite{reddy2021dnsmos}. Due to space constraints, a detailed discussion of this metric is provided in the supplementary material.

\noindent \textbf{Uni-modal Audio Quality - SpeechQual\&Nat.} 
Assesses overall speech quality and naturalness, again restricting this metric to the human linguistic subset. For implementation, we derive a single Mean Opinion Score (MOS) prediction, leveraging NISQAv2~\cite{mittag2021nisqa} (mos-only weights).

\noindent \textbf{Uni-modal Audio Quality - AudioAesthetic.}
Assesses audio aesthetic and production value via four key dimensions: Content Enjoyment (CE), Content Usefulness (CU), Production Complexity (PC), and Production Quality (PQ). We utilize the Audiobox~\cite{tjandra2025meta} module for implementation. Inspired by~\cite{shan2025hunyuanvideo}, which suggests PC is inversely correlated with perceived quality, we define our aggregated score (higher is better) as: \begin{equation} \text{S}_{\text{audioaesthetic}} = \frac{\text{CE} + \text{CU} + \text{PQ} - \text{PC}}{4} \end{equation}

\noindent \textbf{Cross-modal Semantic Alignment - Text-Video Align.} Evaluates semantic consistency between generated videos and input text prompts. Given the complex temporal information in videos, we select ViCLIP~\cite{wang2023internvid}, specifically designed for video understanding to more accurately capture dynamic correspondences.

\noindent \textbf{Cross-modal Semantic Alignment - Text-Audio Align.} Assesses semantic consistency between generated audio and input text prompts. We utilize the CLAP~\cite{wu2023large} model to calculate the cosine similarity between audio and text embeddings.

\noindent \textbf{Cross-modal Semantic Alignment - Audio-Visual Align.} Measures semantic matching between visual frames and audio track. We leverage ImageBind's~\cite{girdhar2023imagebind} robust joint embedding space to calculate the similarity between visual and audio embeddings.

\noindent \textbf{Temporal Synchronization - Desync.} Assesses fine-grained temporal alignment between audio and visual streams. For implementation, we compute the predicted desynchronization offset using Synchformer~\cite{iashin2024synchformer}. Inspired by MMAudio~\cite{cheng2025mmaudio}, we analyze the first and last 4.8s of the video (allowing overlap) to assess alignment.

\noindent \textbf{Temporal Synchronization - Lip-Sync.} Assesses the synchronization between lip movements and speech, a crucial metric for talking heads. This metric is applied exclusively to the human linguistic subset and only where a talking head is detected. For implementation, we calculate the alignment confidence, inspired by the evaluation method of LatentSync~\cite{li2024latentsync}.

\subsubsection{MLLM-based Evaluation}

While traditional human-based Mean Opinion Scores (MOS) for video quality are labor-intensive, unscalable, and subjective, emerging Omni-modal LLMs~\cite{xu2025qwen2} offer an efficient, standardized alternative. Our framework leverages these models for evaluation at two complementary levels: coarse-grained (macro, scored 1-5) and fine-grained (micro).Specific implementation details are provided in the supplementary material.

\noindent \textbf{Macro - Alignment.} Assesses audio-visual coherence, assessing both temporal synchronization (alignment of audio-visual events) and semantic correspondence (natural, logical coordination of multimodal content).

\noindent \textbf{Macro - Artistry.} Assesses the aesthetic and expressive quality of the audio-visual fusion, distinct from technical realism. It evaluates the stylistic unity, creative intent, and resulting artistic impact of the synergy. Due to space constraints, a detailed discussion of this metric is provided in the supplementary material.

\noindent \textbf{Macro - Expressiveness.} Assesses audio's storytelling effectiveness, evaluating how its emotional alignment reinforces the intended mood and its narrative function clarifies or enhances the on-screen story.

\noindent \textbf{Macro - Audio Realism.} Assesses the physical plausibility of the audio track, examining if attributes like loudness and timbre conform to real-world acoustic laws. This metric is computed excluding the virtual worlds category.

\noindent \textbf{Macro - Visual Realism.} Assesses the physical plausibility of the video frames, examining if aspects like lighting and motion fluency adhere to real-world physical laws. This metric is computed excluding the virtual worlds category.

\noindent \textbf{Micro - Audio QA Pairs.}
Designed to assess generation quality from a fine-grained acoustic perspective. For each video sample, a set of 3 to 7 questions is designed to focus on the physical properties of sound and details of environmental interactions.

\noindent \textbf{Micro - Visual QA Pairs.}
Designed to assess generation quality from a fine-grained visual perspective. Similarly, a specific question set (3-7 questions) is tailored for each sample, focusing on visual elements and physical action details mentioned in the input text prompts. 

\vspace{0.5em}

In the micro-level evaluation, for a test set containing $N$ video samples, let $K_i$ be the number of detail-oriented questions generated for the $i$-th sample, and let $C_i$ denote the number of those questions for which the LLM judge determines that the video sample satisfies the corresponding detail requirement. The final fine-grained score $S$ is defined as the average accuracy across all samples:
\begin{equation}
    S = \frac{1}{N} \sum_{i=1}^{N} \frac{C_i}{K_i}
\end{equation}

\begin{table*}[ht]
\centering
\caption{T2AV evaluation results. The results for AV and V+A models are separated by a horizontal line. Underlined scores indicate the highest within each category (AV or V+A), and bolded scores indicate the overall best for each metric.}
\resizebox{\textwidth}{!}{%
\begin{tabular}{c|c|c|c|c|c|c|c|c|c|c|c|c|c}
\toprule[1pt]
\textbf{Models} &
\makecell{\textbf{Speech}\\ \textbf{Q\&N}} &
\makecell{\textbf{Audio}\\ \textbf{Aes}} &
\makecell{\textbf{T-V} \\ \textbf{Align}} &
\makecell{\textbf{T-A} \\ \textbf{Align}} &
\makecell{\textbf{A-V} \\ \textbf{Align}} &
\makecell{\textbf{Lip-} \\ \textbf{Sync}} &
% \makecell{\textbf{De-} \\ \textbf{sync}$\downarrow$} &
\textbf{Desync}$\downarrow$ &
\makecell{\textbf{Align-} \\ \textbf{ment}} &
\makecell{\textbf{Expres-} \\ \textbf{siveness}} &
\makecell{\textbf{Visual} \\ \textbf{Realism}} &
\makecell{\textbf{Audio} \\ \textbf{Realism}} &
\makecell{\textbf{Audio} \\ \textbf{QA}} &
\makecell{\textbf{Visual} \\ \textbf{QA}} \\
\Xhline{0.7pt}

sora2  & 2.672 & 2.867 & 0.2256 & 0.3465 & 0.2376 & 2.655 & 0.7167 & 4.546  & 4.379 & \underline{\textbf{4.805}} & \underline{\textbf{4.375}} & \underline{\textbf{0.8082}} & 0.7994 \\

veo3  & \underline{\textbf{3.073}} & \underline{\textbf{3.543}} & \underline{\textbf{0.2304}} & \underline{\textbf{0.3582}} & \underline{\textbf{0.3164}} & 3.294 & 0.5184 & \underline{\textbf{4.553}} & 4.424 & 4.773 & 4.309 & 0.7999 & \underline{\textbf{0.8095}} \\

wan2.5  & 2.562 & 3.061 & 0.2275 & 0.3033 & 0.2099 & \underline{\textbf{3.671}} & \underline{\textbf{0.4622}} & 4.465 & \underline{\textbf{4.441}} & 4.674 & 4.185 & 0.7993 & 0.7933 \\
\hline

seed\_think  & 2.274 & 2.793 & \multirow{2}{*}{0.2215} & 0.2730 & 0.1960 & \underline{2.785} & 0.5730 & 4.459 & 4.338 & \multirow{2}{*}{\underline{4.724}} & 4.159 & 0.6698 & \multirow{2}{*}{0.7265} \\

seed\_mm  & 2.352 & 2.900 &  & 0.3365 & 0.2817 & 1.743 & \underline{0.4792} & \underline{4.506} & 4.354 &  & 4.175 & 0.6741 &  \\

wan2.2\_think  & 2.116 & 2.825 & \multirow{2}{*}{0.2128} & 0.2735 & 0.2090 & 1.559 & 0.6049 & 4.279 & 4.310 & \multirow{2}{*}{4.649} & 4.069 & 0.5647 & \multirow{2}{*}{0.6363} \\

wan2.2\_mm  & 2.159 & 2.839 &  & 0.3385 & 0.2775 & 1.401 & 0.5429 & 4.377 & 4.318 &  & 4.109 & 0.5861 &  \\

kling\_think  & 2.369 & 2.901 & \multirow{2}{*}{\underline{\textbf{0.2304}}} & 0.2692 & 0.2203 & 2.144 & 0.5946 & 4.455 & \underline{4.409} & \multirow{2}{*}{4.720} & 4.182 & 0.6932 & \multirow{2}{*}{\underline{0.7754}} \\

kling\_mm  & \underline{2.465} & \underline{2.954} &  & \underline{0.3500} & \underline{0.2929} & 1.740 & 0.5617 & 4.440 & 4.408 &  & \underline{4.197} & \underline{0.7300} &  \\
\bottomrule[1pt]
\end{tabular}%
}
\label{tab:t2va_all}
\end{table*}

\begin{table*}[ht]
\caption{I2AV evaluation results, following the same presentation protocol as \cref{tab:t2va_all} (T2AV). }
\resizebox{\textwidth}{!}{%
\begin{tabular}{c|c|c|c|c|c|c|c|c|c|c|c}
\toprule[1pt]
\textbf{Models} &
\makecell{\textbf{Audio}\\ \textbf{Aes}} &
\makecell{\textbf{T-V} \\ \textbf{Align}} &
\makecell{\textbf{T-A} \\ \textbf{Align}} &
\makecell{\textbf{A-V} \\ \textbf{Align}} &
\textbf{Desync$\downarrow$} & 
\textbf{Alignment} & 
\makecell{\textbf{Expres-} \\ \textbf{siveness}} &
\makecell{\textbf{Visual} \\ \textbf{Realsim}} &
\makecell{\textbf{Audio} \\ \textbf{Realism}} &
\textbf{Audio QA} & 
\textbf{Visual QA} \\
\Xhline{0.7pt}
sora2 & 2.974 & 0.2188 & 0.4045 & 0.2623 & 0.9171 & 4.885 & 4.390 & \underline{\textbf{4.964}} & 4.597 & 0.8287 & 0.7611 \\
veo3 & \underline{\textbf{3.574}} & 0.2334 & \underline{\textbf{0.4130}} & \underline{\textbf{0.3215}} & 0.6136 & \underline{4.906} & \underline{4.631} & 4.921 & \underline{4.660} & \underline{\textbf{0.8584}} & \underline{\textbf{0.7982}} \\
wan2.5 & 3.455 & \underline{\textbf{0.2374}} & 0.2865 & 0.2112 & \underline{\textbf{0.3539}} & 4.812 & 4.495 & 4.766 & 4.395 & 0.8084 & 0.7889 \\
\hline
seed\_think & 2.834 & \multirow{2}{*}{0.2276} & 0.3183 & 0.2341 & 0.6845 & 4.879 & 4.528 & \multirow{2}{*}{4.901} & 4.491 & 0.7833 & \multirow{2}{*}{\underline{0.7453}} \\
seed\_mm & \underline{2.974} & & 0.4074 & \underline{0.3185} & \underline{0.5885} & \underline{\textbf{4.918}} & \underline{\textbf{4.585}} & & \underline{\textbf{4.526}} & \underline{0.8020} & \\
wan2.2\_think & 2.833 & \multirow{2}{*}{\underline{0.2292}} & 0.3216 & 0.2145 & 0.7977 & 4.846 & 4.503 & \multirow{2}{*}{\underline{4.913}} & 4.392 & 0.7334 & \multirow{2}{*}{0.7338} \\
wan2.2\_mm & 2.933 & & \underline{0.4121} & 0.3053 & 0.7225 & 4.898 & 4.516 & & 4.420 & 0.7553 & \\
kling\_think & 2.882 & \multirow{2}{*}{0.2270} & 0.3169 & 0.2310 & 0.7102 & 4.860 & 4.507 & \multirow{2}{*}{4.893} & 4.369 & 0.7425 & \multirow{2}{*}{0.7418} \\
kling\_mm & 2.948 & & 0.4052 & 0.3128 & 0.6772 & 4.879 & 4.522 & & 4.424 & 0.7918 & \\
\bottomrule[1pt]
\end{tabular}%
}
\label{tab:i2va_all}
\end{table*}

\subsubsection{Stereophonic Analysis}
We also evaluated the stereophonic performance of generated audio based on human check and nine core acoustic metrics, categorized into two primary dimensions: Spatial Imaging Quality and Signal Integrity \& Compatibility.

\noindent \textbf{Spatial Imaging Quality.} This dimension assesses spatial distribution and sound image clarity. We measure stereo width (Mid/Side energy ratio~\cite{pulkki2015communication}), imaging stability (fluctuations in ITD~\cite{rspa1909}), level stability (fluctuations in ILD~\cite{rspa1909}), and inter-channel temporal consistency (envelope correlation and transient synchronization~\cite{Kates2002}).

\noindent \textbf{Signal Integrity \& Compatibility.} This dimension focuses on technical stability and cross-device compatibility. We calculate phase coherence across low, mid, and high frequency bands~\cite{Kates2002}. We also assess mono downmixing fidelity using the mono compatibility metric (mono loss percent)~\cite{thiede2000peaq} and its inverted form, the Mono Compat score (defined as 1 - normalized mono loss). For directional consistency, we apply inverse normalization to Mono Compat, Imaging Stability, and Level Stability. Finally, we present a nine-dimensional radar chart to visualize and quantify model performance across spatial imaging and signal integrity for stereophonic generation.

\section{Experiment}
\label{sec:exp}

\subsection{Implement}

We evaluated T2AV and I2AV tasks on VABench using two system categories. The first is end-to-end AV models: Veo3-fast, Wan2.5 Preview, and Sora2. The second is decoupled V+A models: combinations of a video generator (Seedance-1.0-lite~\cite{gao2025seedance}, Wan2.2-TI2V~\cite{wan2025wan}, Kling2.5 Turbo~\cite{kling20252.5turbo}) and an audio model (ThinkSound light, MMAudio). Veo3-fast, Wan2.5 Preview, Sora2, Seedance-1.0-lite, Kling2.5 Turbo were accessed via APIs. ThinkSound light, Wan2.2-TI2V, MMAudio were deployed locally. Video outputs were set to 720P (or closest aspect ratio; Wan2.2/2.5 adjust automatically), with frame rate and duration following default settings. For audio, 48kHz stereo tracks were extracted from AV models, while V+A models' outputs (ThinkSound, MMAudio) were retained in their native output formats. Prompts for audio models were modified following official guidelines to satisfy the input length limits while maintaining core semantics.

\begin{figure}[t]
    \centering
    \includegraphics[width=0.95\linewidth]
    {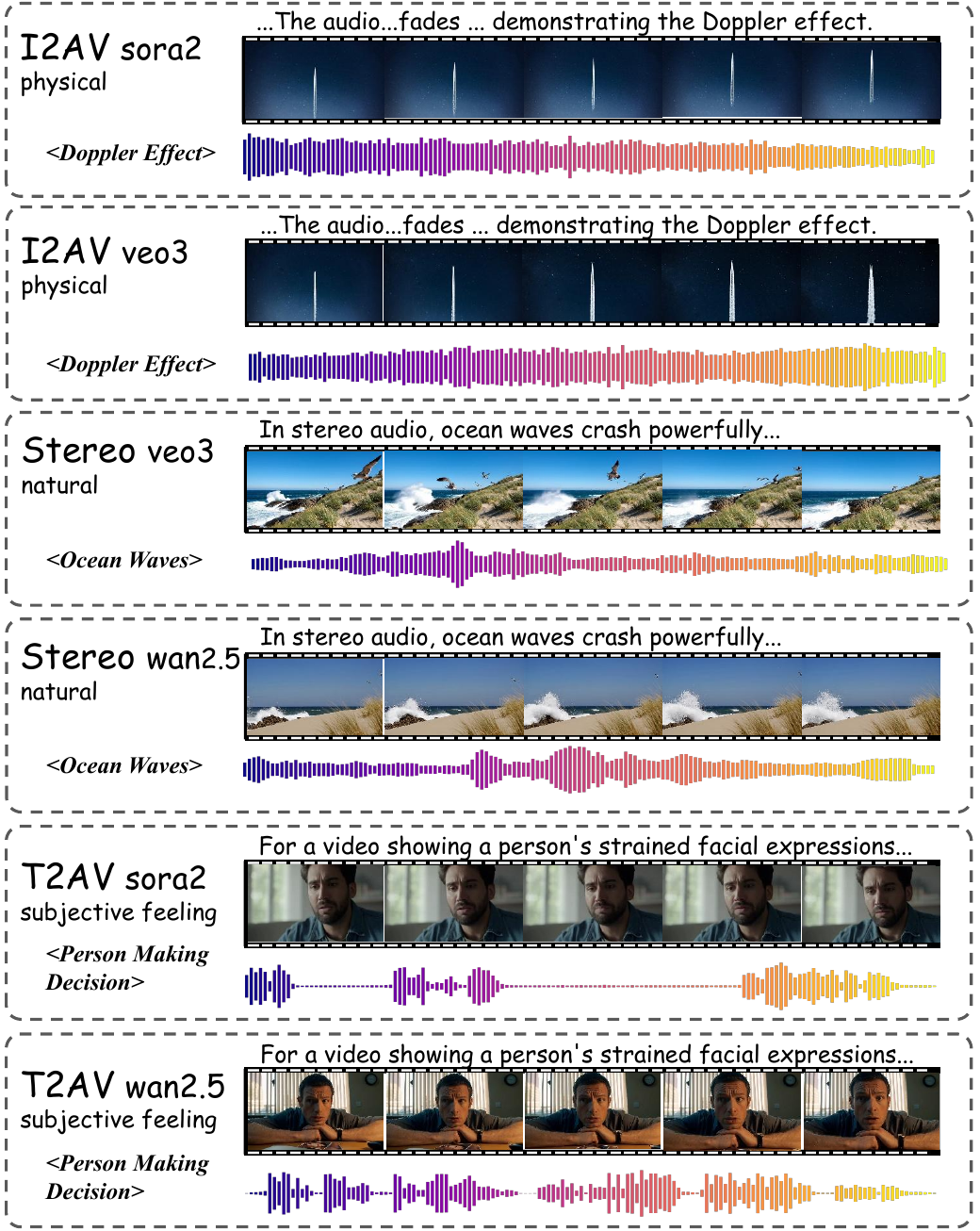}
    \vspace{-0.3em}
    \caption{Qualitative comparison of model performance. We visualize pairwise comparisons across three tasks (I2AV, Stereo, T2AV) by showing key video frames and audio waveforms.}
    \label{fig:qual_compare}
    % \vspace{-4em}
\end{figure}

In all tables and figures, we use unified abbreviations. AV models: veo3 (Veo3-Fast), wan2.5 (Wan2.5 Preview), sora2 (Sora2). V+A models: prefixes seed (Seedance-1.0-Lite), wan2.2 (Wan2.2 TI2V), kling (Kling2.5-Turbo) denote the video generator, while suffixes mm (MMAudio), think (ThinkSound light) denote the audio model, with the two parts joined by an underscore (\_). We use simplified names (e.g., Seedance) in the main text for clarity. Generated results are shown in \cref{fig:qual_compare}.

\subsection{Main Results}

\textbf{Text to Audio-Video Generation.}
As shown in \cref{tab:t2va_all}, among AV models, Veo3 demonstrates the strongest overall performance, particularly in audio quality and cross-modal semantic alignment. Sora2 excels in realism but lags in audio aesthetics and synchronization. Wan2.5 achieves the best audio-visual synchronization, especially Lip-Sync, though its semantic alignment is slightly lower than Veo3 and Sora2. The results of the three models indicate that semantic consistency, synchronization, and realism are difficult to achieve simultaneously.

Among audio models, MMAudio generally outperforms ThinkSound, while ThinkSound shows advantages in Lip-Sync. For pure visual models, Kling achieves the highest visual quality across nearly all vision metrics. Seedance, while slightly weaker overall, unlocks the potential for superior Lip-Sync from the audio model. Kling+MMAudio, combining the strongest audio and video models, stands out as the strongest V+A model, indicating that higher-quality video generation can facilitate improved audio generation.

Overall, integrated AV models tend to hold an advantage, suggesting that end-to-end joint training more effectively captures cross-modal synergies, forms a unified semantic space, and generates more natural, coherent audio-visual content. Nonetheless, the V+A approach remains a viable alternative.

% --------------------------------------------- %

\noindent \textbf{Image to Audio-Video Generation.}
As shown in \cref{tab:i2va_all}, the overall performance of the three AV models mirrors that of T2AV. Among pure visual models, Seedance achieves the strongest results, ranking first across nearly all metrics, while Wan2.2 and Kling perform slightly lower. Seedance+MMAudio attains the best results on major metrics. Overall, integrated AV models generally outperform V+A models, highlighting the effectiveness of end-to-end joint training.

\noindent \textbf{T2AV and I2AV joint analysis.}
Compared with T2AV, the stronger visual constraints brought by input images in I2AV reduce performance gaps among models. Some V+A combinations (e.g., Seedance+MMAudio) even surpass AV models in Alignment, though AV models retain a clear advantage in T2AV. For Expressiveness, the gap between AV and V+A models narrows. For fine-grained semantics (QA), AV models lead in both tasks, but visual input reduces differences and improves score balance. Overall, the results of T2AV and I2AV show consistent trends with stable variations across evaluation dimensions.

\begin{figure}[t]
    \centering
    \begin{subfigure}{\linewidth}
        \centering
        \includegraphics[width=0.95\linewidth]{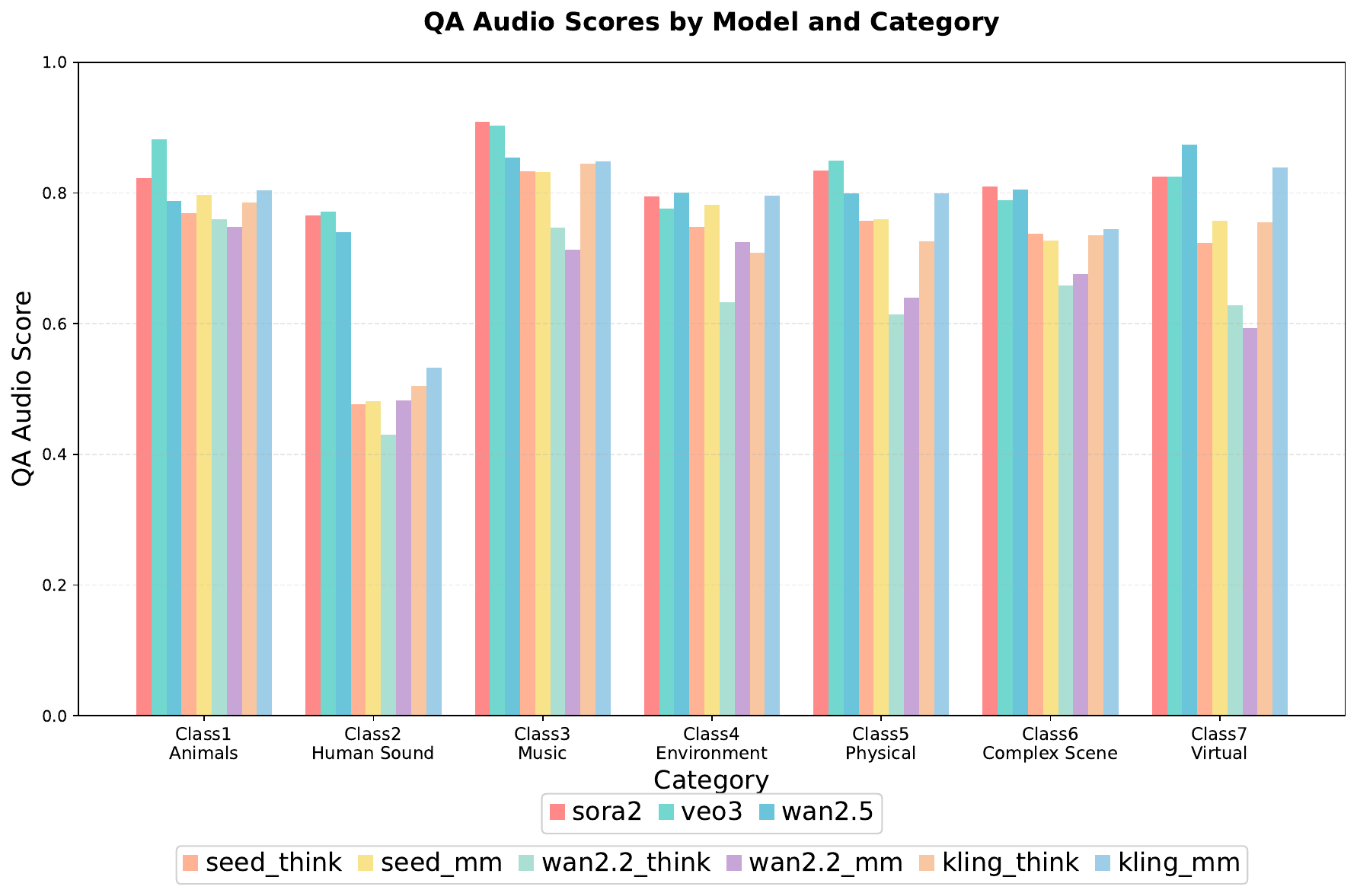}
        \caption{Audio QA}
        \label{fig:qa_aud}
    \end{subfigure}\\[3pt] % 调整垂直间距
    \begin{subfigure}{\linewidth}
        \centering
        \includegraphics[width=0.95\linewidth]{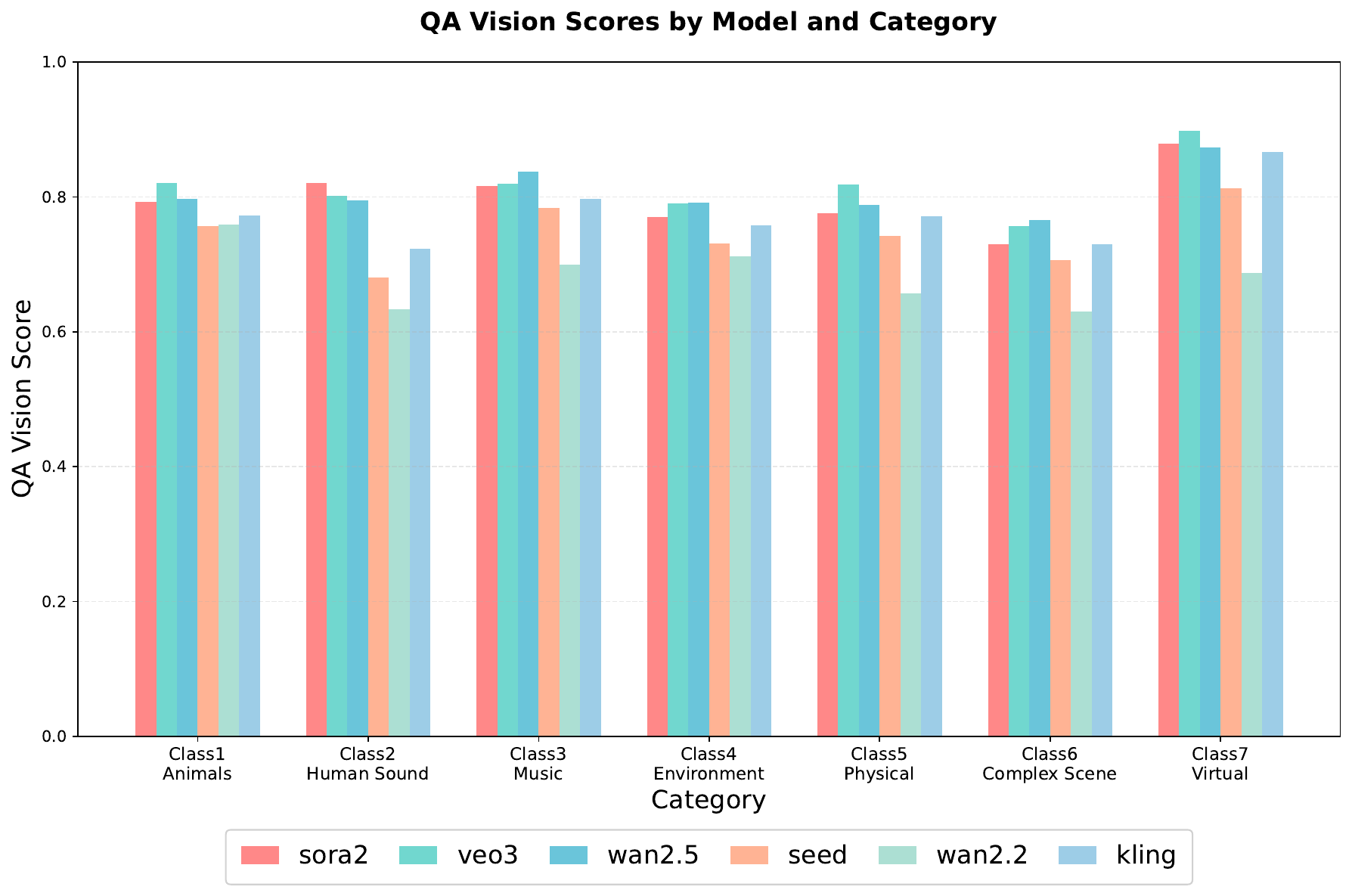}
        \caption{Vision QA}
        \label{fig:qa_vid}
    \end{subfigure}
    \vspace{-1.8em}
    \caption{Fine-grained QA evaluation across seven audio categories for different model architectures.}
    \label{fig:qa_whole}
\end{figure}

\subsection{Additional Analysis}

\textbf{Multi-Categories Analysis.}
The result of AQA is shown in \cref{fig:qa_aud}. Veo3 achieves the strongest overall performance among AV models, excelling in Animals, while Sora2 delivers the most balanced results. Wan2.5 performs best in the Virtual Worlds category.

Within the V+A framework, a clear performance hierarchy emerges: Kling ranks first, followed by Seedance, with Wan2.2 trailing. On the audio side, MMAudio performs robustly overall, whereas ThinkSound demonstrates distinct advantages in Music.

Overall, current systems perform well on weakly correlated audio types such as Music and Animals but struggle with Human Sounds. For highly synchronized tasks, AV models—benefiting from end-to-end joint modeling—consistently outperform V+A ones. Notably, even the best V+A combination fails to surpass the weakest AV model, while all three AV models exhibit small performance gaps and stable results, reflecting their technical maturity and architectural advantages.

% ------------------------------ 下面是qa_video%
The VQA results (\cref{fig:qa_vid}) follow a similar trend as the AQA analysis. Among AV models, Veo3 delivers the strongest overall performance, excelling in Animals, Synchronous Physical Sounds, and Virtual Worlds. Sora2 ranks first in Human Sounds, while Wan2.5 performs robustly in Music and Complex Scenes. The distribution of pure visual models mirrors the AQA results.

Models perform best in the Virtual Worlds category, whereas Complex Scenes receives the lowest scores, revealing persistent challenges in multi-source dynamic interactions. Notably, AV models dominate the top positions across all categories, with the largest gaps over V+A models observed in Human Sounds and Virtual Worlds. These results highlight that integrating audio cues—spatial, material, and rhythmic—enhances not only visual understanding but also emotional tone and expressiveness, underscoring the systemic advantage of unified audio-visual modeling.

\begin{figure}[t]
    \centering
    \includegraphics[width=0.85\linewidth]{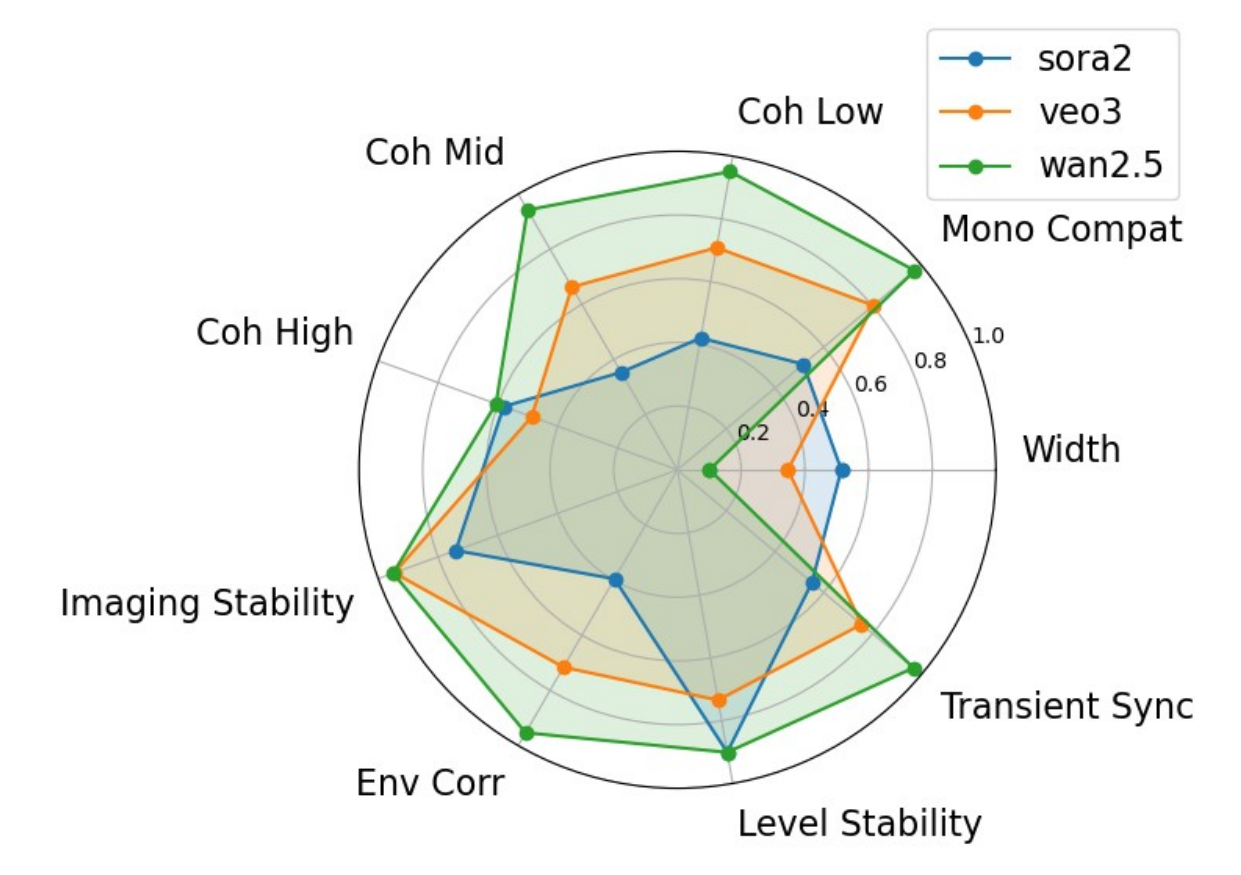}
    \vspace{-0.8em}
    \caption{Comparative radar chart of three models: Phase Coherence (Coh Low/Mid/High), Mono Compatibility (Mono Compat), Soundstage Width (Width), Transient Synchronization (Transient Sync), Level Stability, Envelope Correlation (Env Corr), and Imaging Stability. Higher values indicate better performance.}
    \label{fig:stereo_an}
\end{figure}

\noindent \textbf{Stereo Audio Video Generation.}
Based on nine normalized acoustic metrics (\cref{fig:stereo_an}), the three VA models exhibit a clear trade-off between spatial width and signal fidelity. Wan2.5 demonstrates the best technical fidelity, with highly consistent left–right channels in both time and amplitude domains, but presents the narrowest soundstage. Sora2 features the widest spatial field and stable level balance, yet its width mainly arises from inter-channel phase offsets, leading to unstable localization and energy loss. Veo3 maintains a balance between the two, achieving the most stable sound image and a natural stereo structure.

Human evaluation confirms that none reliably generate stereo separation from text prompts. Sora2 shows noticeable channel-level differences in loudness without semantic distinction; Wan2.5 is nearly monophonic; Veo3 occasionally produces subtle inter-channel alternation and depth cues in natural scenes (e.g., waves, thunder). In certain samples, Veo3 generates moving spatial sources aligned with visual motion, while Sora2 produces distinct left–right vocal tracks in multi-speaker scenes.

Overall, current models lack consistent stereo generation, while localized spatialization in Veo3 and Sora2 implies spatial audio cues in their training data, informing future research on spatial hearing in audio-video generation.

\begin{figure}[t]
    \centering
    \includegraphics[width=0.88\linewidth]{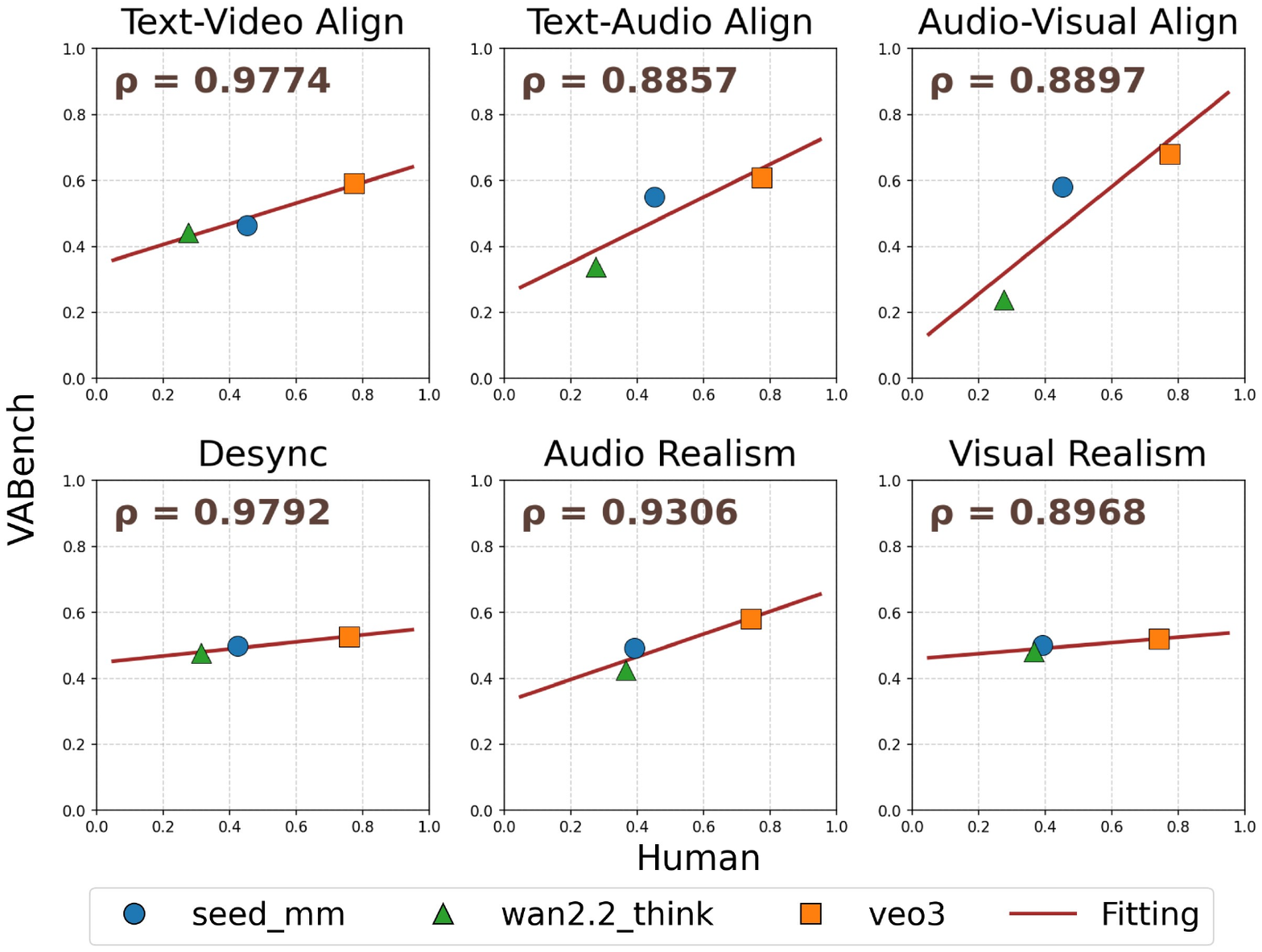}
    \vspace{-0.4em}
    \caption{Human preference consistency validation. Each subplot shows one evaluation dimension, where each point denotes a model’s win rate (x: human, y: VABench). A reference line indicates their correlation, with the Pearson coefficient ($\rho$) annotated.}
    \label{fig:user_stdy}
\end{figure}

\subsection{User Study}

To validate VABench's alignment with human senses, we conducted a pilot user study. Balancing evaluation fidelity and cost, we had six professional evaluators rate a representative subset of videos from three models (Veo3, Seedance+MMAudio, Wan2.2+ThinkSound) on a 1–5 scale. The evaluation focused on three key dimensions—semantics, synchronization, and realism—which correspond to aggregated benchmark metrics (Semantics: Text-Video Align, Text-Audio Align, Audio-Visual Align; Synchronization: Desync; Realism: Audio Realism, Video Realism). We then computed pairwise win rates (Win=1, Loss=0, Tie=0.5) for human ratings and benchmark scores, averaging all comparisons to get per-model win rates. Finally, we calculated the Pearson correlation between human- and benchmark-derived win rates. The results, shown in \cref{fig:user_stdy}, demonstrate a strong correlation between VABench and human preferences across dimensions.

\section{Conclusion}
\label{sec:conclusion}

We present VABench, a comprehensive benchmark for evaluating synchronous audio-video generation across T2AV, I2AV, and stereo tasks. Featuring automated, multidimensional, and human-aligned evaluation, VABench enables reliable and interpretable performance assessment while revealing the challenge of balancing semantics, synchronization, and realism.
We believe that VABench provides valuable insights for achieving more coherent and perceptually grounded audio-video generation, and will serve as a robust contribution to research and evaluation in this field.

\noindent \textbf{Acknowledgments:}

This work is supported by National Natural Science Foundation of China (92470121, 62402016), National Key R\&D Program of China (2024YFA1014003), Zhongguancun Academy (C20250204, C20250602),  Beijing Major Science and Technology Project (Z251100008125043, Z251100008425023), and High-performance Computing Platform of Peking University. This work is also supported by Ant Group through CCF-Ant Research Fund.

{
    \small
    \bibliographystyle{ieeenat_fullname}
    \bibliography{main}
}

% \clearpage
\maketitlesupplementary
\setcounter{page}{1}

%-----------------------两个额外的指标分析
\section{Additional evaluation metrics}

\subsection{Supplementry results analysis in SpeechClarity and Artistry}

\begin{table}[ht]
\caption{Supplementary results for T2AV and I2AV}
\label{tab:av_supp}
\centering
\begin{tabular}{c|cc|c}
\toprule[1pt]
\multirow{2}{*}{\textbf{Models}} & \multicolumn{2}{c|}{\textbf{T2AV}} & \textbf{I2AV} \\ \cline{2-4} 
 & \textbf{SpeechClarity} & \textbf{Artisry} & \textbf{Artisry} \\
\Xhline{0.7pt}
sora2 & 2.367 & 3.735 & 3.931 \\
veo3 & \underline{\textbf{2.554}} & 3.825 & \underline{\textbf{3.983}} \\
wan2.5 & 2.396 & \underline{3.838} & 3.929 \\
\hline
seed\_think & 2.008 & 3.717 & 3.956 \\
seed\_mm & 2.202 & 3.707 & \underline{3.971} \\
wan2.2\_think & 1.882 & 3.630 & 3.942 \\
wan2.2\_mm & 2.016 & 3.609 & 3.962 \\
kling\_think & 2.051 & \underline{\textbf{3.844}} & 3.950 \\
kling\_mm & \underline{2.221} & \underline{\textbf{3.844}} & 3.958 \\
\bottomrule[1pt]
\end{tabular}
\end{table}

In this part, we present supplementary metrics excluded from the main text(\cref{tab:av_supp}), along with further analysis of these results. Additionally, we extend and validate the primary conclusions of our study based on these supplementary findings.

On the SpeechClarity metric(leveraging DNSMOS~\cite{reddy2021dnsmos}, the AV models collectively achieve the best overall performance, which aligns with our previous analysis: AV models significantly outperform V+A models in representing human language, thereby enabling more comprehensive optimization of speech quality. For the V+A approach, Kling~\cite{kling20252.5turbo}, Seedance~\cite{gao2025seedance}, and Wan2.2~\cite{wan2025wan} exhibit a descending performance trend, reaffirming that higher-quality visual generation substantially enhances the latent capabilities of audio models. On the Artistry metric(leveraging Qwen2.5 Omni 7B~\cite{xu2025qwen2}, AV models maintain an overall lead, yet Kling + MMAudio~\cite{cheng2025mmaudio} reaches the current state-of-the-art level.

Beyond the primary conclusions presented in the main text, our extended evaluation across comprehensive metrics reveals several representative performance differentiations. Specifically, Veo3~\cite{google2025veo3} demonstrates precise synergistic control over acoustic and visual details within its joint modeling framework, while Sora2~\cite{openai2025sora2} exhibits more prominent capabilities in Synchronous Physical Sounds (hereafter referred to as Physical)) plausibility and event consistency. Among visual-only models, Kling shows superior performance in both artistic style and visual fidelity compared to peer approaches. Furthermore, Kling+MMAudio surpasses Sora2 on cross-modal metrics such as Text-Audio Align and Audio-Visual Align, while also demonstrating robust performance across subjective dimensions including Artistry, Expressiveness, and Audio Realism. These supplementary observations provide nuanced substantiation for the main conclusions.

\subsection{Category Analysis}
This part presents a more complete exposition of the core findings from the Multi-Categories Analysis in Section 4.3, supplemented by additional discoveries.

\noindent \textbf{Audio QA.}
The three AV models demonstrate robust performance with distinct specializations. Veo3 leads in Animals but shows relative weakness in Environmental Sounds (hereafter referred to as Environment) and Complex Scenes categories. Sora2 delivers the most balanced performance, consistently ranking within the top two, which highlights its strong generalization capability. Wan2.5~\cite{alibaba2025wan25} dominates the Virtual Worlds (hereafter referred to as Virtual) category yet exhibits noticeable shortcomings in Music, Physical, and Animals. Notably, Human Sounds remains the most challenging domain for all models, indicating a shared limitation in simulating human vocal signals.

Within the V+A architecture group, MMAudio exhibits comprehensive capabilities, showing particular strength in the Environment and Virtual categories, where its performance approaches that of top-tier AV models. Meanwhile, ThinkSound demonstrates aspecialization in the Music category, highlighting its specific proficiency in musical generation.

Comparing Wan2.2 and Wan2.5, both models exhibit a performance gap relative to peers in the Physical category, suggesting that modeling physical laws remains a challenging area for this model family. A similar trend is observed in the Music category, indicating that these specific semantic domains may benefit from further optimization.

Among video generation models, Kling achieves the highest performance in the Human Sounds and Virtual categories. Seedance occupies the middle tier, marginally surpassing Kling in the Environment category, while Wan2.2's performance trails in Music, Physical, and Virtual. Experimental results confirm that high-quality video inputs can significantly augment audio generation potential, a synergy clearly evident in the performance gains of the Kling + MMAudio combination.

Overall, the consistent superiority of AV models underscores the architectural advantage of end-to-end joint training in achieving high-fidelity, tightly-coupled generation. This advantage is particularly pronounced in the Human Sounds category, where precise spatiotemporal synchronization is paramount. These findings not only delineate specific deficiencies but also inform targeted optimization strategies; for instance, V+A architectures could be significantly enhanced by strengthening speech generation modules or integrating specialized vocal models to address current limitations in human sounds synthesis.

\noindent\textbf{Visual QA.}

The evaluation reveals distinct capability specializations among the three AV models. Veo3 distinguishes itself in scenarios demanding sophisticated physical logic and complex dynamics. Sora2 demonstrates superior proficiency in human-centric modeling. Meanwhile, Wan2.5 exhibits exceptional performance in multi-element, non-biological environments.

Among pure visual models, Kling surpasses certain V+A models in the most challenging Complex Scenes category, demonstrating the significant potential of top-tier visual-only generation frameworks.In contrast, Wan2.2 trails behind in multiple critical categories (e.g., Physical, Virtual, and Music), suggesting that complex scene comprehension and cross-element consistency remain challenging aspects for its architecture.

The results underscore the structural advantages of AV models, which consistently occupy the top three positions across all categories. Beyond multimodal alignment, the integration of audio signals enhances holistic scene understanding; spatial cues, material properties, and event dynamics provide critical context, contributing to more physically plausible and temporally coherent visuals. This benefit is most critical in the Human Sounds category, where the millisecond-level precision required for lip synchronization leverages the joint architecture to address alignment challenges that remain significant for pure visual frameworks.

Further analysis suggests that audio is critical for visual generation, extending beyond temporal synchronization. This is evidenced by the Virtual category, which exhibits the second-largest performance gap between model types. In abstract or surreal scenarios, auditory cues—such as rhythm, energy distribution, and emotional tone—provide essential structural guidance. AV models leverage these signals to enhance visual dynamism and narrative coherence, whereas pure visual models face the increased challenge of inferring these attributes without multimodal guidance.

These results demonstrate that even in the most creative, free-generation domains, end-to-end audio-visual co-design significantly enhances both generation quality and creative consistency.

\begin{table}[!ht]
\centering
\caption{The attributes of the videos generated by each model.}
\label{tab:video_attr}
\resizebox{0.55\columnwidth}{!}{%
\begin{tabular}{c|cc}
\toprule
Models & Length & FPS \\ \hline
sora2 & 10s & 30 \\
veo3 & 8s & 24 \\
wan2.5 & 5s & 24 \\
seedance 1.0 lite & 5s & 24 \\
wan2.2 & 5s & 24 \\
kling2.5 turbo & 5s & 24 \\
\bottomrule
\end{tabular}%
}
\end{table}

\begin{table*}[!ht]
\centering
\caption{Results of different models on the T2AV task across seven audio categories, including \texttt{Virtual}. For each model, the highest score within each metric is highlighted in bold, and the lowest score is denoted with ‘\textsuperscript{$\dagger$}’.}
\resizebox{\textwidth}{!}{%
\begin{tabular}{c|c|c|c|c|c|c|c|c|c|c|c|c|c}
\toprule[1pt]
\textbf{Model} &
\textbf{Category} &
\makecell{\textbf{Audio-} \\ \textbf{Aes}} &
\makecell{\textbf{T-V} \\ \textbf{Align}} &
\makecell{\textbf{T-A} \\ \textbf{Align}} &
\makecell{\textbf{A-V} \\ \textbf{Align}} &
\textbf{Desync$\downarrow$} &
\makecell{\textbf{Align-} \\ \textbf{ment}} &
\textbf{Artistry} &
\makecell{\textbf{Expres-} \\ \textbf{siveness}} &
\makecell{\textbf{Visual} \\ \textbf{Realism}} &
\makecell{\textbf{Audio} \\ \textbf{Realism}} &
\makecell{\textbf{Audio} \\ \textbf{QA}} &
\makecell{\textbf{Visual} \\ \textbf{QA}} \\
\Xhline{0.7pt}

sora2 & Animals & 3.309 & \textbf{0.2647} & 0.3428 & 0.2923 & 0.7943 & 4.914 & 3.714 & \textbf{4.657} & \textbf{4.971} & 4.629 & 0.8300 & \textbf{0.9023} \\
sora2 & Human Sounds & 3.047 & 0.2179\textsuperscript{$\dagger$} & 0.2848\textsuperscript{$\dagger$} & 0.2308 & 0.6072 & 4.383 & 3.689 & 4.367 & 4.794 & 4.406 & 0.7597\textsuperscript{$\dagger$} & 0.8197 \\
sora2 & Music & \textbf{3.512} & 0.2300 & \textbf{0.5521} & \textbf{0.3328} & \textbf{0.4710} & \textbf{4.968} & 3.968 & 4.613 & 4.968 & \textbf{4.903} & \textbf{0.9091} & 0.7660 \\
sora2 & Environment & 2.509\textsuperscript{$\dagger$} & 0.2345 & 0.4008 & 0.2531 & 0.8205 & 4.795 & 3.671 & 4.233 & 4.890 & 4.575 & 0.8057 & 0.7867 \\
sora2 & Physical & 2.842 & 0.2193 & 0.3458 & 0.2229\textsuperscript{$\dagger$} & 0.7342 & 4.592 & 3.418\textsuperscript{$\dagger$} & 4.168\textsuperscript{$\dagger$} & 4.875 & 4.332 & 0.8177 & 0.7849 \\
sora2 & Complex Scene & 2.758 & 0.2225 & 0.3432 & 0.2313 & 0.7344 & 4.344\textsuperscript{$\dagger$} & 3.896 & 4.432 & 4.650 & 4.169 & 0.8173 & 0.7453\textsuperscript{$\dagger$} \\
sora2 & Virtual & 2.675 & 0.2361 & 0.3621 & 0.2278 & 0.8315\textsuperscript{$\dagger$} & 4.696 & \textbf{4.120} & 4.652 & 3.087\textsuperscript{$\dagger$} & 2.674\textsuperscript{$\dagger$} & 0.8255 & 0.8786 \\
\midrule

veo3 & Animals & 3.546 & \textbf{0.2617} & 0.3434 & 0.3692 & 0.8200\textsuperscript{$\dagger$} & 4.914 & 3.829 & 4.429 & 4.914 & 4.200 & 0.8424 & 0.8796 \\
veo3 & Human Sounds & 3.669 & 0.2157\textsuperscript{$\dagger$} & 0.3254\textsuperscript{$\dagger$} & 0.3280 & 0.3794 & 4.383\textsuperscript{$\dagger$} & 3.717 & 4.350 & 4.717 & 4.322 & 0.7668\textsuperscript{$\dagger$} & 0.7985 \\
veo3 & Music & \textbf{4.402} & 0.2345 & \textbf{0.5319} & \textbf{0.3795} & \textbf{0.3032} & \textbf{4.968} & 4.064 & 4.581 & \textbf{4.968} & \textbf{4.936} & \textbf{0.8668} & 0.7395\textsuperscript{$\dagger$} \\
veo3 & Environment & 3.298\textsuperscript{$\dagger$} & 0.2430 & 0.3988 & 0.2986 & 0.7589 & 4.740 & 3.795 & 4.260\textsuperscript{$\dagger$} & 4.863 & 4.438 & 0.7873 & 0.7965 \\
veo3 & Physical & 3.411 & 0.2291 & 0.3826 & 0.3097 & 0.3880 & 4.565 & 3.543\textsuperscript{$\dagger$} & 4.342 & 4.793 & 4.310 & 0.8382 & 0.8327 \\
veo3 & Complex Scene & 3.464 & 0.2238 & 0.3287 & 0.3214 & 0.6197 & 4.432 & 3.934 & 4.481 & 4.710 & 4.159 & 0.7669 & 0.7560 \\
veo3 & Virtual & 3.620 & 0.2512 & 0.3470 & 0.2694\textsuperscript{$\dagger$} & 0.6163 & 4.674 & \textbf{4.326} & \textbf{4.696} & 2.891\textsuperscript{$\dagger$} & 2.707\textsuperscript{$\dagger$} & 0.8247 & \textbf{0.8981} \\
\midrule

wan2.5 & Animals & 3.375 & \textbf{0.2545} & 0.3702 & 0.2534 & 0.4286 & 4.343 & 3.829 & 4.600 & \textbf{4.857} & 4.029 & 0.7914 & 0.8676 \\
wan2.5 & Human Sounds & 3.265 & 0.2161\textsuperscript{$\dagger$} & 0.2495\textsuperscript{$\dagger$} & 0.2201 & \textbf{0.2233} & 4.289\textsuperscript{$\dagger$} & 3.733 & 4.344 & 4.667 & 4.178 & 0.7437\textsuperscript{$\dagger$} & 0.7909 \\
wan2.5 & Music & \textbf{3.785} & 0.2194 & \textbf{0.4122} & \textbf{0.2692} & 0.9742\textsuperscript{$\dagger$} & \textbf{4.806} & 4.064 & \textbf{4.645} & 4.806 & \textbf{4.806} & 0.8421 & 0.7582\textsuperscript{$\dagger$} \\
wan2.5 & Environment & 2.950 & 0.2368 & 0.3255 & 0.2034 & 0.5205 & 4.753 & 3.795 & 4.288\textsuperscript{$\dagger$} & 4.822 & 4.356 & 0.8116 & 0.7822 \\
wan2.5 & Physical & 3.013 & 0.2234 & 0.3112 & 0.2126 & 0.4500 & 4.571 & 3.625\textsuperscript{$\dagger$} & 4.402 & 4.685 & 4.228 & 0.7689 & 0.8020 \\
wan2.5 & Complex Scene & 2.898 & 0.2246 & 0.3160 & 0.2080 & 0.5607 & 4.301 & 3.967 & 4.497 & 4.552 & 4.005 & 0.8110 & 0.7686 \\
wan2.5 & Virtual & 2.803\textsuperscript{$\dagger$} & 0.2491 & 0.2880 & 0.1570\textsuperscript{$\dagger$} & 0.5522 & 4.630 & \textbf{4.174} & 4.587 & 2.848\textsuperscript{$\dagger$} & 2.663\textsuperscript{$\dagger$} & \textbf{0.8739} & \textbf{0.8732} \\
\midrule

seed\_think & Animals & 2.996 & \textbf{0.2519} & 0.3259 & 0.2433 & 0.7543 & 4.571 & 3.629 & 4.543 & 4.829 & 3.829 & 0.6667 & 0.7995 \\
seed\_think & Human Sounds & 2.689 & 0.2025\textsuperscript{$\dagger$} & 0.2194 & 0.1869 & 0.4678 & 4.150\textsuperscript{$\dagger$} & 3.533 & 4.150\textsuperscript{$\dagger$} & 4.700 & 4.139 & 0.4635\textsuperscript{$\dagger$} & 0.6735\textsuperscript{$\dagger$} \\
seed\_think & Music & \textbf{3.998} & 0.2234 & \textbf{0.4561} & \textbf{0.3187} & \textbf{0.2194} & \textbf{4.936} & 3.968 & \textbf{4.548} & \textbf{5.000} & \textbf{4.903} & \textbf{0.8281} & 0.7069 \\
seed\_think & Environment & 2.501\textsuperscript{$\dagger$} & 0.2378 & 0.3101 & 0.1859 & 0.7452 & 4.753 & 3.753 & 4.329 & 4.842 & 4.425 & 0.7661 & 0.7377 \\
seed\_think & Physical & 2.892 & 0.2250 & 0.2941 & 0.1966 & 0.4913 & 4.462 & 3.467\textsuperscript{$\dagger$} & 4.293 & 4.796 & 4.163 & 0.7269 & 0.7436 \\
seed\_think & Complex Scene & 2.688 & 0.2121 & 0.2760 & 0.1934 & 0.6098 & 4.448 & 3.863 & 4.404 & 4.560 & 4.005 & 0.7232 & 0.7030 \\
seed\_think & Virtual & 2.752 & 0.2450 & 0.2189\textsuperscript{$\dagger$} & 0.1664\textsuperscript{$\dagger$} & 0.7826\textsuperscript{$\dagger$} & 4.641 & \textbf{4.207} & 4.522 & 2.761\textsuperscript{$\dagger$} & 2.413\textsuperscript{$\dagger$} & 0.7241 & \textbf{0.8124} \\
\midrule

seed\_mm & Animals & 3.288 & \textbf{0.2519} & 0.3127 & 0.3453 & 0.6457 & 4.629 & 3.600 & 4.429 & 4.829 & 4.000 & 0.7133 & 0.7995 \\
seed\_mm & Human Sounds & 2.922 & 0.2025\textsuperscript{$\dagger$} & 0.2709\textsuperscript{$\dagger$} & 0.2743 & 0.3689 & 4.178\textsuperscript{$\dagger$} & 3.561 & 4.156\textsuperscript{$\dagger$} & 4.700 & 4.128 & 0.4645\textsuperscript{$\dagger$} & 0.6735\textsuperscript{$\dagger$} \\
seed\_mm & Music & \textbf{3.814} & 0.2237 & \textbf{0.4772} & \textbf{0.4252} & \textbf{0.3097} & \textbf{5.000} & 3.968 & \textbf{4.581} & \textbf{5.000} & \textbf{4.936} & \textbf{0.8198} & 0.7069 \\
seed\_mm & Environment & 2.618\textsuperscript{$\dagger$} & 0.2378 & 0.4224 & 0.3158 & 0.7699\textsuperscript{$\dagger$} & 4.781 & 3.726 & 4.315 & 4.842 & 4.438 & 0.7868 & 0.7377 \\
seed\_mm & Physical & 2.940 & 0.2250 & 0.3720 & 0.2722 & 0.4087 & 4.652 & 3.391\textsuperscript{$\dagger$} & 4.348 & 4.796 & 4.212 & 0.7233 & 0.7436 \\
seed\_mm & Complex Scene & 2.755 & 0.2121 & 0.3354 & 0.2804 & 0.4667 & 4.377 & 3.885 & 4.404 & 4.560 & 3.984 & 0.7113 & 0.7030 \\
seed\_mm & Virtual & 2.832 & 0.2450 & 0.2894 & 0.2178\textsuperscript{$\dagger$} & 0.6239 & 4.685 & \textbf{4.207} & 4.576 & 2.761\textsuperscript{$\dagger$} & 2.511\textsuperscript{$\dagger$} & 0.7581 & \textbf{0.8124} \\
\midrule

wan2.2\_think & Animals & 2.902 & \textbf{0.2542} & 0.3324 & 0.2630 & 0.8171 & 4.257 & 3.457 & 4.371 & 4.500 & 3.743 & \textbf{0.7105} & \textbf{0.8322} \\
wan2.2\_think & Human Sounds & 2.682 & 0.1963\textsuperscript{$\dagger$} & 0.2227\textsuperscript{$\dagger$} & 0.2074 & 0.5011 & 3.922\textsuperscript{$\dagger$} & 3.511 & 4.172\textsuperscript{$\dagger$} & 4.597 & 4.089 & 0.4175\textsuperscript{$\dagger$} & 0.6220 \\
wan2.2\_think & Music & \textbf{4.053} & 0.2144 & \textbf{0.4134} & \textbf{0.3332} & \textbf{0.2387} & \textbf{4.774} & 3.839 & \textbf{4.452} & 4.832 & \textbf{4.677} & 0.6458 & 0.5422\textsuperscript{$\dagger$} \\
wan2.2\_think & Environment & 2.524\textsuperscript{$\dagger$} & 0.2287 & 0.3297 & 0.1975 & 0.8603 & 4.726 & 3.671 & 4.315 & \textbf{4.884} & 4.315 & 0.6375 & 0.6974 \\
wan2.2\_think & Physical & 3.033 & 0.2133 & 0.2879 & 0.2019 & 0.4533 & 4.370 & 3.332\textsuperscript{$\dagger$} & 4.239 & 4.736 & 4.005 & 0.5323 & 0.6284 \\
wan2.2\_think & Complex Scene & 2.689 & 0.2074 & 0.2730 & 0.2102 & 0.6372 & 4.186 & 3.858 & 4.432 & 4.530 & 3.973 & 0.6395 & 0.5870 \\
wan2.2\_think & Virtual & 2.749 & 0.2258 & 0.2306 & 0.1707\textsuperscript{$\dagger$} & 0.8870\textsuperscript{$\dagger$} & 4.467 & \textbf{3.967} & 4.402 & 3.103\textsuperscript{$\dagger$} & 2.717\textsuperscript{$\dagger$} & 0.6280 & 0.6872 \\
\midrule

wan2.2\_mm & Animals & 3.121 & \textbf{0.2542} & 0.3152 & 0.3452 & 0.7257 & 4.429 & 3.486 & 4.486 & 4.500 & 3.800 & \textbf{0.6905} & \textbf{0.8322} \\
wan2.2\_mm & Human Sounds & 2.848 & 0.1963\textsuperscript{$\dagger$} & 0.2887\textsuperscript{$\dagger$} & 0.2820 & 0.4800 & 4.011\textsuperscript{$\dagger$} & 3.472 & 4.161\textsuperscript{$\dagger$} & 4.597 & 4.072 & 0.4653\textsuperscript{$\dagger$} & 0.6220 \\
wan2.2\_mm & Music & \textbf{3.743} & 0.2144 & \textbf{0.4505} & \textbf{0.3903} & \textbf{0.2903} & \textbf{4.806} & 3.839 & \textbf{4.581} & 4.823 & \textbf{4.677} & 0.5988 & 0.5422\textsuperscript{$\dagger$} \\
wan2.2\_mm & Environment & 2.570\textsuperscript{$\dagger$} & 0.2287 & 0.4167 & 0.2916 & 0.7068 & 4.575 & 3.712 & 4.288 & \textbf{4.884} & 4.411 & 0.6894 & 0.6974 \\
wan2.2\_mm & Physical & 2.948 & 0.2133 & 0.3740 & 0.2651 & 0.946 & 4.522 & 3.288\textsuperscript{$\dagger$} & 4.261 & 4.736 & 4.141 & 0.5794 & 0.6284 \\
wan2.2\_mm & Complex Scene & 2.674 & 0.2074 & 0.3259 & 0.2865 & 0.5770 & 4.333 & 3.869 & 4.410 & 4.530 & 3.956 & 0.6446 & 0.5870 \\
wan2.2\_mm & Virtual & 2.733 & 0.2258 & 0.2989 & 0.2008\textsuperscript{$\dagger$} & 0.7804\textsuperscript{$\dagger$} & 4.565 & \textbf{3.891} & 4.424 & 3.103\textsuperscript{$\dagger$} & 2.717\textsuperscript{$\dagger$} & 0.5937 & 0.6872 \\
\midrule

kling\_think & Animals & 2.963 & \textbf{0.2555} & 0.3122 & 0.2670 & 0.5600 & 4.743 & 3.800 & 4.486 & 4.857 & 4.114 & 0.8286 & \textbf{0.8828} \\
kling\_think & Human Sounds & 2.701\textsuperscript{$\dagger$} & 0.2103\textsuperscript{$\dagger$} & 0.2344\textsuperscript{$\dagger$} & 0.2026\textsuperscript{$\dagger$} & 0.5900 & 4.250\textsuperscript{$\dagger$} & 3.694 & 4.244\textsuperscript{$\dagger$} & 4.708 & 4.172 & 0.4937\textsuperscript{$\dagger$} & 0.7171\textsuperscript{$\dagger$} \\
kling\_think & Music & \textbf{4.101} & 0.2236 & \textbf{0.4374} & \textbf{0.3502} & \textbf{0.2258} & \textbf{4.968} & 4.032 & \textbf{4.710} & \textbf{4.935} & \textbf{4.871} & \textbf{0.9012} & 0.7408 \\
kling\_think & Environment & 2.813 & 0.2363 & 0.2911 & 0.2046 & 0.6904 & 4.863 & 3.781 & 4.397 & 4.884 & 4.384 & 0.7807 & 0.7980 \\
kling\_think & Physical & 3.135 & 0.2339 & 0.2730 & 0.2114 & 0.3989 & 4.380 & 3.636\textsuperscript{$\dagger$} & 4.332 & 4.812 & 4.217 & 0.7035 & 0.7950 \\
kling\_think & Complex Scene & 2.672 & 0.2304 & 0.2658 & 0.2239 & 0.7410 & 4.322 & 4.000 & 4.470 & 4.508 & 3.973 & 0.7519 & 0.7436 \\
kling\_think & Virtual & 2.920 & 0.2505 & 0.2456 & 0.2166 & 0.7652\textsuperscript{$\dagger$} & 4.663 & \textbf{4.250} & 4.641 & 2.989\textsuperscript{$\dagger$} & 2.772\textsuperscript{$\dagger$} & 0.7548 & 0.8662 \\
\midrule

kling\_mm & Animals & 3.139 & \textbf{0.2555} & 0.3446 & 0.3491 & 0.6917 & 4.543 & 3.800 & 4.514 & 4.857 & 4.057 & 0.7691 & \textbf{0.8828} \\
kling\_mm & Human Sounds & 3.073 & 0.2103\textsuperscript{$\dagger$} & 0.3037 & 0.2827 & 0.5300 & 4.144\textsuperscript{$\dagger$} & 3.661 & 4.244\textsuperscript{$\dagger$} & 4.708 & 4.183 & 0.5203\textsuperscript{$\dagger$} & 0.7171\textsuperscript{$\dagger$} \\
kling\_mm & Music & \textbf{4.011} & 0.2236 & \textbf{0.4937} & \textbf{0.4082} & \textbf{0.2581} & \textbf{4.968} & 4.032 & 4.548 & \textbf{4.935} & \textbf{4.903} & \textbf{0.8991} & 0.7408 \\
kling\_mm & Environment & 2.582\textsuperscript{$\dagger$} & 0.2362 & 0.4279 & 0.2988 & 0.6575 & 4.836 & 3.767 & 4.370 & 4.884 & 4.384 & 0.8347 & 0.7980 \\
kling\_mm & Physical & 3.049 & 0.2339 & 0.3766 & 0.2952 & 0.4120 & 4.489 & 3.658\textsuperscript{$\dagger$} & 4.370 & 4.812 & 4.207 & 0.7910 & 0.7950 \\
kling\_mm & Complex Scene & 2.722 & 0.2304 & 0.3386 & 0.2920 & 0.6033 & 4.339 & 4.016 & 4.475 & 4.508 & 4.033 & 0.7420 & 0.7436 \\
kling\_mm & Virtual & 2.859 & 0.2505 & 0.3019\textsuperscript{$\dagger$} & 0.2453\textsuperscript{$\dagger$} & 0.8152\textsuperscript{$\dagger$} & 4.587 & \textbf{4.250} & \textbf{4.609} & 2.989\textsuperscript{$\dagger$} & 2.707\textsuperscript{$\dagger$} & 0.8398 & 0.8662 \\
\bottomrule[1pt]
\end{tabular}%
}
\label{tab:catres_t2av}
\end{table*}

\begin{table*}[!ht]
\centering
\caption{Results of different models on the I2AV task across seven audio categories, excluding \texttt{Virtual}. Same notation as \cref{tab:catres_t2av}.}
\resizebox{\textwidth}{!}{%
\begin{tabular}{c|c|c|c|c|c|c|c|c|c|c|c|c|c}
\toprule[1pt]
\textbf{Model} &
\textbf{Category} &
\makecell{\textbf{Audio-} \\ \textbf{Aes}} &
\makecell{\textbf{T-V} \\ \textbf{Align}} &
\makecell{\textbf{T-A} \\ \textbf{Align}} &
\makecell{\textbf{A-V} \\ \textbf{Align}} &
\textbf{Desync$\downarrow$} &
\makecell{\textbf{Align-} \\ \textbf{ment}} &
\textbf{Artistry} &
\makecell{\textbf{Expres-} \\ \textbf{siveness}} &
\makecell{\textbf{Visual} \\ \textbf{Realism}} &
\makecell{\textbf{Audio} \\ \textbf{Realism}} &
\makecell{\textbf{Audio} \\ \textbf{QA}} &
\makecell{\textbf{Visual} \\ \textbf{QA}} \\
\Xhline{0.7pt}

sora2 & Animals & 3.164 & \textbf{0.2321} & 0.4556 & \textbf{0.2952} & 0.9277 & 4.830 & 3.921 & \textbf{4.525} & 4.949 & \textbf{4.627} & 0.8210 & 0.7708 \\
sora2 & Human Sounds & 2.536\textsuperscript{$\dagger$} & 0.2233 & 0.3870 & 0.1997\textsuperscript{$\dagger$} & 1.140\textsuperscript{$\dagger$} & 4.600\textsuperscript{$\dagger$} & \textbf{4.000} & 4.200\textsuperscript{$\dagger$} & 4.800\textsuperscript{$\dagger$} & 4.500 & 0.8667 & 0.8405 \\
sora2 & Music & \textbf{3.880} & 0.1934\textsuperscript{$\dagger$} & \textbf{0.4709} & 0.2618 & \textbf{0.5180} & \textbf{5.000} & \textbf{4.000} & 4.420 & 4.980 & \textbf{4.860} & \textbf{0.9100} & \textbf{0.8474} \\
sora2 & Environment & 2.642 & 0.2154 & 0.3420\textsuperscript{$\dagger$} & 0.2585 & 1.095 & 4.888 & 3.950 & 4.287 & \textbf{5.000} & 4.487\textsuperscript{$\dagger$} & 0.7855\textsuperscript{$\dagger$} & 0.7540 \\
sora2 & Physical & 2.850 & 0.2142 & 0.3782 & 0.2356 & 0.7346 & 4.923 & 3.875\textsuperscript{$\dagger$} & 4.279 & 4.981 & 4.548 & 0.8636 & 0.7596 \\
sora2 & Complex Scene & 2.622 & 0.2150 & 0.3646 & 0.2412 & 1.123 & 4.910 & 3.950 & 4.350 & 4.950 & 4.560 & 0.7960 & 0.6998\textsuperscript{$\dagger$} \\
\midrule

veo3 & Animals & 3.573 & \textbf{0.2498} & \textbf{0.4884} & \textbf{0.3665} & 0.4785 & 4.830 & 3.966 & \textbf{4.780} & 4.927 & 4.678 & 0.8904 & 0.8083 \\
veo3 & Human Sounds & 3.245\textsuperscript{$\dagger$} & 0.2347 & 0.3661 & 0.2689\textsuperscript{$\dagger$} & 0.6200 & 4.700\textsuperscript{$\dagger$} & 4.000 & 4.300\textsuperscript{$\dagger$} & 4.700\textsuperscript{$\dagger$} & 4.400\textsuperscript{$\dagger$} & 0.8650 & 0.8467 \\
veo3 & Music & \textbf{4.651} & 0.2022\textsuperscript{$\dagger$} & 0.4683 & 0.3612 & \textbf{0.3660} & \textbf{5.000} & 4.000 & 4.680 & \textbf{4.980} & \textbf{4.860} & \textbf{0.9263} & \textbf{0.8696} \\
veo3 & Environment & 3.461 & 0.2318 & 0.3030\textsuperscript{$\dagger$} & 0.2905 & 0.9025\textsuperscript{$\dagger$} & 4.925 & 3.987 & 4.537 & 4.963 & 4.638 & 0.7653\textsuperscript{$\dagger$} & 0.7841 \\
veo3 & Physical & 3.255 & 0.2265 & 0.4131 & 0.2961 & 0.4933 & 4.971 & 3.923\textsuperscript{$\dagger$} & 4.510 & 4.952 & 4.615 & 0.8709 & 0.7926 \\
veo3 & Complex Scene & 3.492 & 0.2284 & 0.3448 & 0.2785 & 0.8700 & 4.930 & \textbf{4.060} & 4.580 & 4.840 & 4.620 & 0.8288 & 0.7568\textsuperscript{$\dagger$} \\
\midrule

wan2.5 & Animals & 3.759 & \textbf{0.2515} & \textbf{0.3229} & \textbf{0.2555} & 0.2768 & 4.746\textsuperscript{$\dagger$} & 3.904 & \textbf{4.667} & 4.706 & 4.378 & 0.7877 & 0.7827 \\
wan2.5 & Human Sounds & 3.120 & 0.2491 & 0.2349 & 0.1400\textsuperscript{$\dagger$} & \textbf{0.1400} & 4.900 & 3.900\textsuperscript{$\dagger$} & 4.200\textsuperscript{$\dagger$} & 4.500\textsuperscript{$\dagger$} & 4.300 & 0.6850\textsuperscript{$\dagger$} & 0.8610 \\
wan2.5 & Music & \textbf{3.877} & 0.2183\textsuperscript{$\dagger$} & 0.2802 & 0.2053 & 0.7040\textsuperscript{$\dagger$} & \textbf{4.980} & \textbf{4.000} & 4.420 & \textbf{4.880} & \textbf{4.700} & \textbf{0.8622} & \textbf{0.8873} \\
wan2.5 & Environment & 3.320 & 0.2321 & 0.2278\textsuperscript{$\dagger$} & 0.1717 & 0.3375 & 4.763 & 3.925 & 4.463 & 4.825 & 4.275\textsuperscript{$\dagger$} & 0.7918 & 0.8005 \\
wan2.5 & Physical & 3.297 & 0.2296 & 0.2922 & 0.2012 & 0.3846 & 4.913 & 3.913 & 4.356 & 4.846 & 4.423 & 0.8546 & 0.7630 \\
wan2.5 & Complex Scene & 3.013\textsuperscript{$\dagger$} & 0.2332 & 0.2714 & 0.1846 & 0.3180 & 4.770 & 3.960 & 4.430 & 4.710 & 4.350 & 0.7958 & 0.7609\textsuperscript{$\dagger$} \\
\midrule

seed\_think & Animals & 2.890 & \textbf{0.2426} & \textbf{0.3789} & 0.2590 & 0.6712 & 4.831 & 3.932 & 4.650 & 4.915 & 4.508 & 0.7895 & 0.7485 \\
seed\_think & Human Sounds & 2.356\textsuperscript{$\dagger$} & 0.2342 & 0.2102\textsuperscript{$\dagger$} & 0.1802\textsuperscript{$\dagger$} & 1.100\textsuperscript{$\dagger$} & 4.400\textsuperscript{$\dagger$} & 3.900\textsuperscript{$\dagger$} & 4.100\textsuperscript{$\dagger$} & 4.650\textsuperscript{$\dagger$} & 4.300\textsuperscript{$\dagger$} & 0.7100\textsuperscript{$\dagger$} & 0.8014 \\
seed\_think & Music & \textbf{3.981} & 0.2035\textsuperscript{$\dagger$} & 0.3691 & \textbf{0.2992} & \textbf{0.2680} & 4.940 & 3.980 & \textbf{4.680} & \textbf{4.950} & \textbf{4.780} & \textbf{0.8363} & \textbf{0.8316} \\
seed\_think & Environment & 2.496 & 0.2218 & 0.2271 & 0.1927 & 0.7225 & 4.888 & 3.938 & 4.463 & 4.906 & 4.375 & 0.7321 & 0.7251 \\
seed\_think & Physical & 2.841 & 0.2231 & 0.2935 & 0.2133 & 0.7077 & \textbf{4.962} & 3.904 & 4.433 & 4.947 & 4.433 & 0.8111 & 0.7405 \\
seed\_think & Complex Scene & 2.469 & 0.2216 & 0.2954 & 0.2179 & 0.8200 & 4.890 & \textbf{4.060} & 4.430 & 4.825 & 4.490 & 0.7651 & 0.7118\textsuperscript{$\dagger$} \\
\midrule

seed\_mm & Animals & 3.117 & \textbf{0.2426} & \textbf{0.4690} & \textbf{0.3572} & 0.5401 & 4.876 & 3.949 & \textbf{4.734} & 4.915 & 4.599 & 0.8145 & 0.7485 \\
seed\_mm & Human Sounds & 2.962 & 0.2342 & 0.2940\textsuperscript{$\dagger$} & 0.2690\textsuperscript{$\dagger$} & 0.7800\textsuperscript{$\dagger$} & 4.700\textsuperscript{$\dagger$} & 4.000 & 4.300\textsuperscript{$\dagger$} & 4.650\textsuperscript{$\dagger$} & 4.200\textsuperscript{$\dagger$} & 0.7833 & 0.8014 \\
seed\_mm & Music & \textbf{4.013} & 0.2035\textsuperscript{$\dagger$} & 0.3608 & 0.3257 & \textbf{0.4880} & \textbf{5.000} & 3.980 & 4.660 & \textbf{4.950} & \textbf{4.820} & \textbf{0.8405} & \textbf{0.8316} \\
seed\_mm & Environment & 2.614\textsuperscript{$\dagger$} & 0.2218 & 0.3610 & 0.2951 & 0.6400 & 4.938 & 4.013 & 4.513 & 4.906 & 4.450 & 0.7789 & 0.7251 \\
seed\_mm & Physical & 2.822 & 0.2231 & 0.4185 & 0.2888 & 0.5173 & 4.952 & 3.923\textsuperscript{$\dagger$} & 4.462 & 4.947 & 4.413 & 0.8257 & 0.7405 \\
seed\_mm & Complex Scene & 2.648 & 0.2216 & 0.3584 & 0.3007 & 0.7380 & 4.920 & \textbf{4.020} & 4.500 & 4.825 & 4.460 & 0.7561\textsuperscript{$\dagger$} & 0.7118\textsuperscript{$\dagger$} \\
\midrule

wan2.2\_think & Animals & 2.837 & \textbf{0.2459} & \textbf{0.3884} & 0.2410 & 0.8102 & 4.808 & 3.949 & \textbf{4.638} & 4.895 & 4.435 & 0.7696 & 0.7440 \\
wan2.2\_think & Human Sounds & 2.641 & 0.2413 & 0.1911\textsuperscript{$\dagger$} & 0.1693\textsuperscript{$\dagger$} & 1.200\textsuperscript{$\dagger$} & 4.400\textsuperscript{$\dagger$} & 3.900 & 4.000\textsuperscript{$\dagger$} & 4.650\textsuperscript{$\dagger$} & 4.100\textsuperscript{$\dagger$} & 0.6667 & \textbf{0.8310} \\
wan2.2\_think & Music & \textbf{4.010} & 0.2008\textsuperscript{$\dagger$} & 0.3629 & \textbf{0.2757} & \textbf{0.2960} & \textbf{4.980} & \textbf{3.980} & 4.620 & 4.940 & \textbf{4.680} & \textbf{0.8091} & 0.7977 \\
wan2.2\_think & Environment & 2.440\textsuperscript{$\dagger$} & 0.2226 & 0.2377 & 0.1764 & 0.9250 & 4.850 & 3.938 & 4.400 & \textbf{4.944} & 4.250 & 0.6289\textsuperscript{$\dagger$} & 0.7248 \\
wan2.2\_think & Physical & 2.946 & 0.2226 & 0.3006 & 0.2082 & 0.6654 & 4.913 & 3.894\textsuperscript{$\dagger$} & 4.394 & 4.942 & 4.337 & 0.7599 & 0.7062\textsuperscript{$\dagger$} \\
wan2.2\_think & Complex Scene & 2.452 & 0.2245 & 0.2846 & 0.1785 & 1.022 & 4.820 & 3.970 & 4.450 & 4.900 & 4.370 & 0.6941 & 0.7097 \\
\midrule

wan2.2\_mm & Animals & 3.015 & \textbf{0.2459} & \textbf{0.4739} & \textbf{0.3462} & 0.7028 & 4.831\textsuperscript{$\dagger$} & 3.932 & \textbf{4.678} & 4.895 & 4.458 & 0.7600 & 0.7440 \\
wan2.2\_mm & Human Sounds & 2.899 & 0.2413 & 0.3111\textsuperscript{$\dagger$} & 0.2729 & 0.8000 & 4.900 & 3.900\textsuperscript{$\dagger$} & 4.300\textsuperscript{$\dagger$} & 4.650\textsuperscript{$\dagger$} & 4.200\textsuperscript{$\dagger$} & \textbf{0.7967} & \textbf{0.8310} \\
wan2.2\_mm & Music & \textbf{3.995} & 0.2008\textsuperscript{$\dagger$} & 0.3552 & 0.3377 & \textbf{0.4160} & \textbf{4.960} & 3.980 & 4.580 & 4.940 & \textbf{4.660} & 0.7842 & 0.7977 \\
wan2.2\_mm & Environment & 2.550\textsuperscript{$\dagger$} & 0.2226 & 0.3759 & 0.2676\textsuperscript{$\dagger$} & 0.7200 & 4.938 & 3.975 & 4.425 & \textbf{4.944} & 4.400 & 0.7583 & 0.7248 \\
wan2.2\_mm & Physical & 2.919 & 0.2226 & 0.4156 & 0.2797 & 0.6808 & 4.952 & 3.952 & 4.375 & 4.942 & 4.317 & 0.7476 & 0.7062\textsuperscript{$\dagger$} \\
wan2.2\_mm & Complex Scene & 2.582 & 0.2245 & 0.3665 & 0.2766 & 0.9480\textsuperscript{$\dagger$} & 4.900 & \textbf{4.010} & 4.440 & 4.900 & 4.380 & 0.7339\textsuperscript{$\dagger$} & 0.7097 \\
\midrule

kling\_think & Animals & 3.009 & \textbf{0.2413} & \textbf{0.3867} & 0.2707 & 0.7311 & 4.791 & 3.915 & \textbf{4.627} & 4.904 & 4.367 & 0.7776 & 0.7503 \\
kling\_think & Human Sounds & 2.431 & 0.2346 & 0.2206 & 0.1351\textsuperscript{$\dagger$} & 1.120\textsuperscript{$\dagger$} & 4.700\textsuperscript{$\dagger$} & 4.000 & 4.100\textsuperscript{$\dagger$} & 4.650\textsuperscript{$\dagger$} & 4.400 & 0.6933 & 0.8240 \\
kling\_think & Music & \textbf{3.980} & 0.2021\textsuperscript{$\dagger$} & 0.3785 & \textbf{0.3128} & \textbf{0.3040} & \textbf{5.000} & \textbf{4.020} & 4.580 & 4.910 & \textbf{4.680} & \textbf{0.8093} & \textbf{0.8313} \\
kling\_think & Environment & 2.439 & 0.2234 & 0.2147\textsuperscript{$\dagger$} & 0.1661 & 0.8150 & 4.862 & 3.987 & 4.425 & \textbf{4.981} & 4.225\textsuperscript{$\dagger$} & 0.6425\textsuperscript{$\dagger$} & 0.7213 \\
kling\_think & Physical & 2.952 & 0.2215 & 0.3020 & 0.2241 & 0.6019 & 4.942 & 3.894\textsuperscript{$\dagger$} & 4.413 & 4.870 & 4.356 & 0.7675 & 0.7287 \\
kling\_think & Complex Scene & 2.433\textsuperscript{$\dagger$} & 0.2216 & 0.2694 & 0.1884 & 0.8640 & 4.840 & 4.000 & 4.460 & 4.840 & 4.340 & 0.7058 & 0.7039\textsuperscript{$\dagger$} \\
\midrule

kling\_mm & Animals & 3.054 & \textbf{0.2413} & \textbf{0.4660} & \textbf{0.3636} & 0.6079 & 4.825 & 3.932 & \textbf{4.706} & 4.904 & 4.503 & 0.8113 & 0.7503 \\
kling\_mm & Human Sounds & 3.196 & 0.2346 & 0.3308\textsuperscript{$\dagger$} & 0.2959 & 1.120\textsuperscript{$\dagger$} & 4.800\textsuperscript{$\dagger$} & 4.000 & 4.500 & 4.650\textsuperscript{$\dagger$} & 4.500 & 0.7567 & 0.8240 \\
kling\_mm & Music & \textbf{3.867} & 0.2021\textsuperscript{$\dagger$} & 0.3701 & 0.3462 & \textbf{0.3400} & \textbf{4.980} & 3.980 & 4.520 & 4.910 & \textbf{4.660} & \textbf{0.8171} & \textbf{0.8313} \\
kling\_mm & Environment & 2.547\textsuperscript{$\dagger$} & 0.2234 & 0.3628 & 0.2719\textsuperscript{$\dagger$} & 0.8325 & 4.925 & 3.975 & 4.412 & \textbf{4.981} & 4.275\textsuperscript{$\dagger$} & 0.7610 & 0.7213 \\
kling\_mm & Physical & 2.895 & 0.2215 & 0.4035 & 0.2739 & 0.6077 & 4.913 & 3.923\textsuperscript{$\dagger$} & 4.375\textsuperscript{$\dagger$} & 4.870 & 4.365 & 0.8136 & 0.7287 \\
kling\_mm & Complex Scene & 2.651 & 0.2216 & 0.3582 & 0.2809 & 0.8720 & 4.860 & \textbf{4.010} & 4.440 & 4.840 & 4.340 & 0.7499\textsuperscript{$\dagger$} & 0.7039\textsuperscript{$\dagger$} \\

\bottomrule[1pt]
\end{tabular}%
}
\label{tab:catres_i2av}
\end{table*}

%---------------------详细的生成参数
\section{Audio-Video Generation Models in Evaluation}
\label{sec:rationale}

% \begin{table}[!ht]
% \centering
% \caption{The attributes of the videos generated by each model.}
% \label{tab:video_attr}
% \resizebox{0.55\columnwidth}{!}{%
% \begin{tabular}{c|cc}
% \toprule
% Models & Length & FPS \\ \hline
% sora2 & 10s & 30 \\
% veo3 & 8s & 24 \\
% wan2.5 & 5s & 24 \\
% seedance 1.0 lite & 5s & 24 \\
% wan2.2 & 5s & 24 \\
% kling2.5 turbo & 5s & 24 \\
% \bottomrule
% \end{tabular}%
% }
% \end{table}

% 名称需要和正文统一一下
In our experiments, we adhered to the default configuration parameters provided by each video generation model, as summarized in \cref{tab:video_attr} . Specifically, these settings include the default output duration (Length) and frame rate (FPS). For instance, Sora2 generates videos of 10 seconds at 30 FPS by default, while Veo3-fast, Wan2.5 Preview, Seedance-1.0-Lite, Wan2.2-TI2V, and Kling2.5 Turbo all produce videos of 5–8 seconds at 24 FPS.
% under their respective default settings.

\section{Detail Analysis of Different Tasks}

This section provides a comprehensive analysis of experimental results across different categories for various models under both T2AV (\cref{tab:catres_t2av}) and I2AV (\cref{tab:catres_i2av}) tasks. The study aims to identify common patterns across tasks and elucidate the specific impact of image-conditioned input (I2AV) on the final outcomes.

\subsection{Common Strengths and Core Challenges}

Despite differences in input modalities, the models' ability to handle specific content categories shows high consistency across both tasks, revealing universal strengths and core bottlenecks in current technologies.

\noindent \textbf{Common Strength: Robustness in the Music Category.}
In both T2AV and I2AV tasks, the Music category yields superior scores across most metrics. This trend suggests that current generative frameworks are particularly adept at processing structured, melodic content. Benefiting from the inherent correlation between musical audio and visual dynamics, models demonstrate sustained stability in achieving high-fidelity synchronization and emotional expression across input modalities.

\noindent \textbf{Common Challenges: Human Sounds and Complex Scenes.}
Human Sounds and Complex Scenes emerge as persistent challenges across tasks, consistently exhibiting lower scores in alignment and macro-evaluation metrics. These categories represent significant technical bottlenecks in current generation frameworks. The difficulty in Human Sounds is attributed to the nuanced and abstract nature of the content (e.g., ``contemplation"), combined with rigorous demands for temporal synchronization regarding character actions, lip movements, and realistic detail. Meanwhile, the Complex Scenes category is constrained by challenges in multi-element interference, multi-source fusion, and comprehensive scene reasoning, resulting in persistent limitations.
Addressing these bottlenecks in human sounds synthesis and complex scene generation remains a priority for future research.

\begin{figure}[t]
    \centering
    \includegraphics[width=1.0\linewidth]{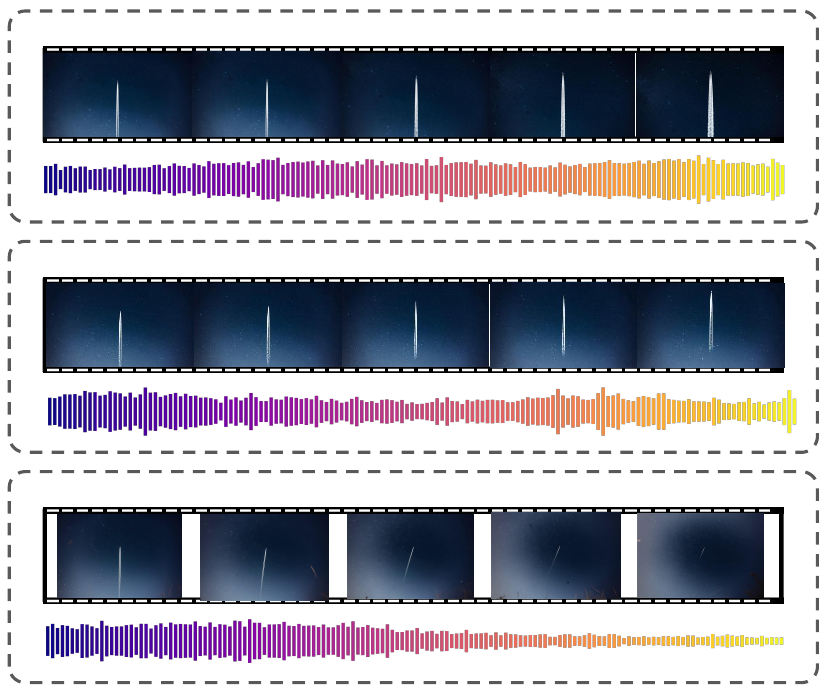}
    \caption{Doppler-effect video for analysis. From top to bottom, the results correspond to the outputs of Veo3, Sora2 and Wan2.5, respectively.}
    \label{fig:doppler_show}
\end{figure}

\subsection{Impact of Image Input}
The use of images as conditional inputs is the key difference between I2AV and T2AV tasks. The experimental results reveal the significant influence of image inputs on the evaluation of generation outcomes.

\noindent \textbf{Convergence and Constraint of Artistry Scores.}
In T2AV tasks, the Virtual category yields the highest Artistry scores, suggesting that models demonstrate peak creative expression when unencumbered by strict physical constraints. Conversely, I2AV tasks exhibit minimal variance in Artistry scores across categories, with values converging toward a central mean (3.8--4.0). This pattern implies that static image inputs impose a constraint on free artistic expression, biasing the generation process toward the physical fidelity and realism inherent in the visual reference.Future methodologies can therefore explore ways to enable high-level artistic expression within these constraints.

\noindent \textbf{Convergence and Stabilization of Alignment and Realism Metrics.}
Compared to T2AV, the inclusion of image inputs in I2AV reduces performance variance across models and raises the minimum performance floor across categories in Alignment, Visual Realism, and Audio Realism. This stabilization effect is particularly pronounced in the Physical category, where explicit visual grounding more effectively constrains physical states and spatial relationships, resulting in markedly more consistent and realistic audio–video generation.

\begin{figure}[t]
    \centering
    \includegraphics[width=1\linewidth]{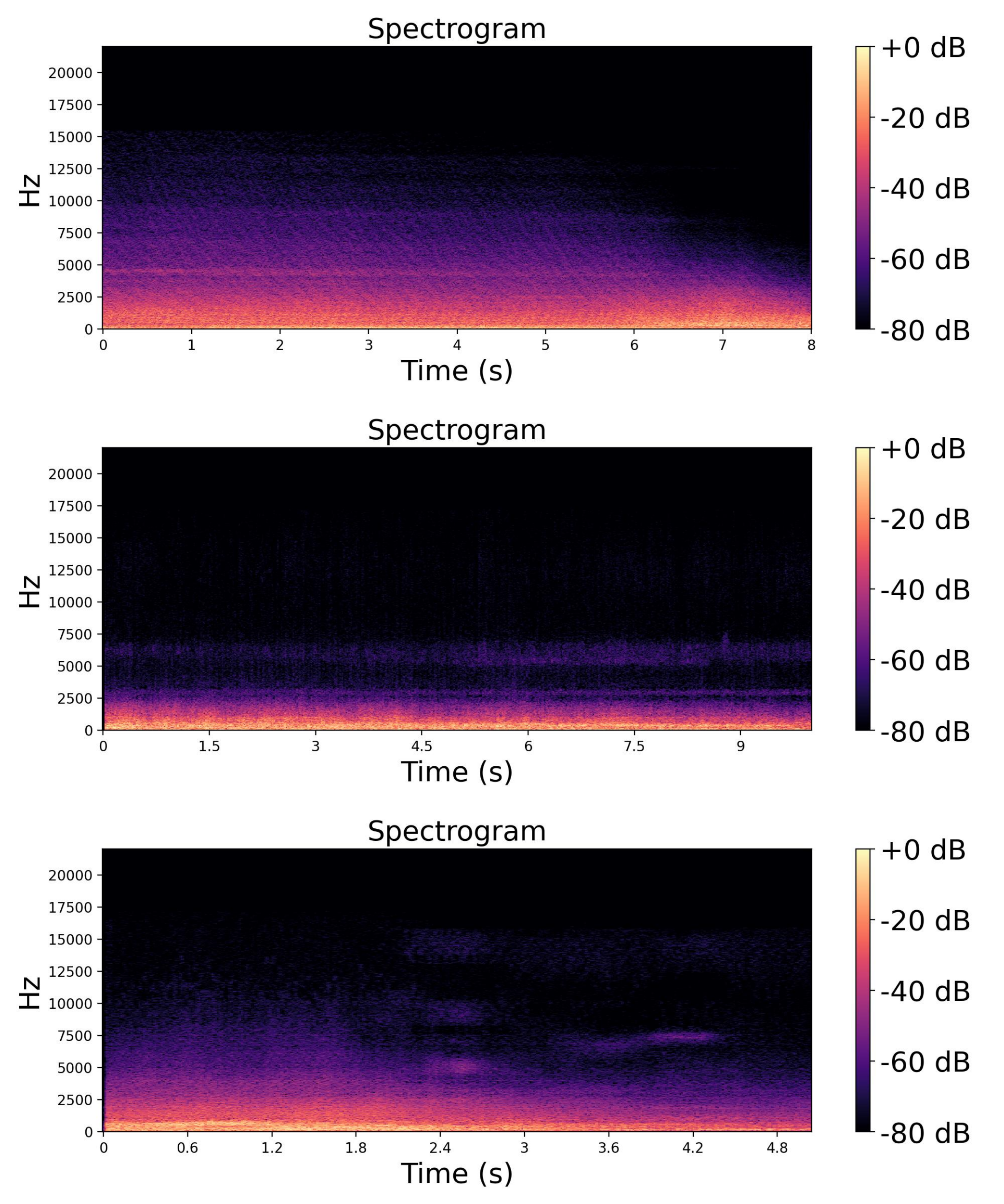}
    \caption{Spectrograms of the video generated by three models. From top to bottom, the results correspond to Veo3, Sora2, and Wan2.5, respectively.}
    \label{fig:doppler_all}
\end{figure}

% ---------------------有意思的样例分析和发现
\section{Qualitative Analysis}
In this section, we conduct a more detailed analysis based on several specific scenarios. These scenarios are selected to examine how the models handle challenging multimodal cues involving physical principles, temporal constraints, and spatial structures.

\begin{figure}[t]
    \centering
    \includegraphics[width=1.0\linewidth]{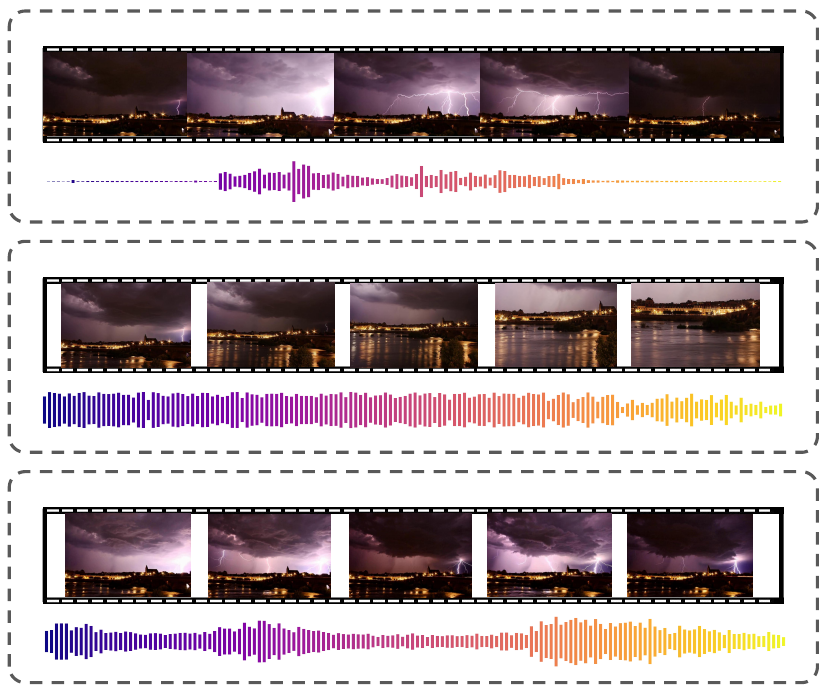}
    \caption{Lightning video for analysis. From top to bottom, the results correspond to the outputs of Veo3, Wan2.5 and Kling+MMAudio, respectively.}
    \label{fig:light_show}
\end{figure}

\subsection{Doppler Effect}

This part evaluates whether the models can generate acoustically plausible variations that conform to the physical principles of the ``approach-pass-recede" dynamic when explicitly prompted for Doppler effect synthesis, thereby authentically reproducing the auditory characteristics of high-speed moving sound sources. For this purpose, we select three AV models and analyze an example (\cref{fig:doppler_show}) through spectrogram visualization of its audio content (see \cref{fig:doppler_all}). The prompt used for this example is as follows:

\begin{quote}
    In the night sky, an airplane flies at high speed, leaving a long trail behind. The audio should include the roaring sound of the airplane engine, which gradually fades as the plane moves away, demonstrating the Doppler effect. Faint wind noise and occasional sounds of nocturnal insects are present in the background, creating a vast auditory atmosphere of the night sky.
\end{quote}

Analysis of the results reveals that Veo3's spectrogram most clearly demonstrates the Doppler effect—its frequency trajectory shows a smooth temporal descent, accurately simulating the pitch variation of an aircraft approaching and receding, while simultaneously rendering the environmental atmosphere specified in the prompt. In comparison, although Wan2.5 captures the gradual attenuation of engine roar with changing distance, its Doppler shift characteristics are less pronounced than Veo3's. As for Sora2, while its Doppler effect is not as prominent as the other two models, the overall auditory perception aligns more closely with human intuition: given the aircraft's altitude and distance in the visual scene, the engine sound should inherently exhibit a lower fundamental frequency accompanied by a moderate degree of frequency shifting, and Sora2 delivers a more perceptually plausible representation in this regard.

\begin{figure}[t]
    \centering
    \includegraphics[width=1.0\linewidth]{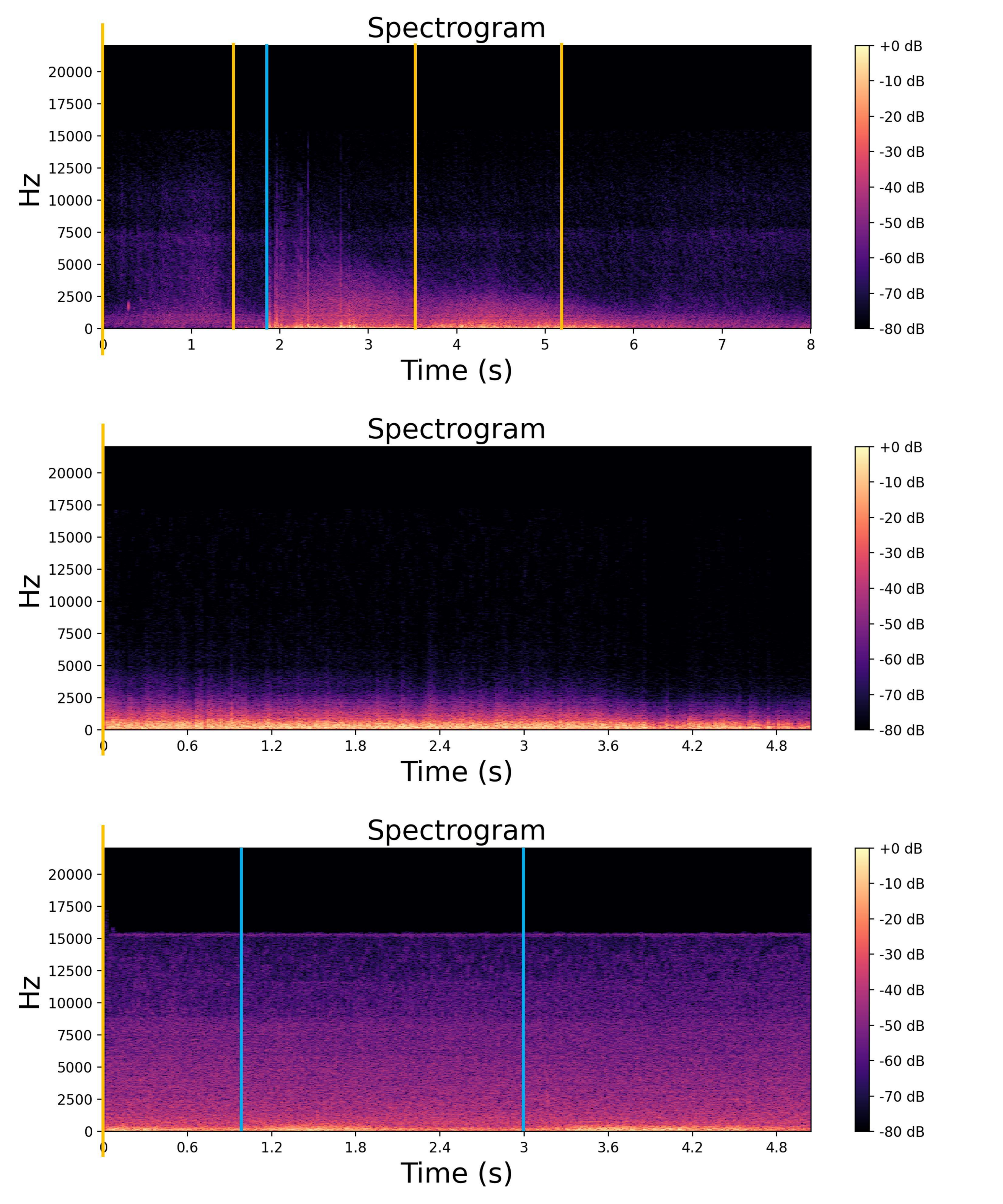}
    \caption{Spectrograms of the lightning video generated by the three models. From top to bottom, the results correspond to Veo3, Wan2.5, and Kling+MMAudio, respectively. Yellow vertical lines mark the approximate timestamps of visible lightning strikes, while blue vertical lines indicate the thunder events (excluding those occurring at the very beginning of the video).}
    \label{fig:light_all}
\end{figure}

\subsection{Lighting}

This section evaluates whether the models can adhere to the natural physical principle of ``thunder following lightning" when generating videos with the prompt ``lightning," thereby producing thunder scenes with physical consistency and perceptual plausibility. We examine three models—Veo3, Wan2.5, and Kling+MMAudio—using a sample (\cref{fig:light_show}) with spectral analysis conducted on the corresponding audio signals (see \cref{fig:light_all}). The prompt used for this example is as follows:

\begin{quote}
    On a pitch-black night, the distant sky is split by lightning, accompanied by the rumbling of thunder. The flash of lightning is brief and bright, while the thunder rolls in from afar, gradually intensifying and shaking the soul. In the surrounding environment, wind howls fiercely, power poles sway slightly in the gusts, and the occasional hum of vibrating wires can be heard. The entire soundscape brims with the power and dynamism of nature, as the low-frequency rumbles of thunder contrast sharply with the high-frequency whistling of the wind.
\end{quote}

From the spectrogram of the Veo3-generated audio, the first thunderclap occurs after lightning is already visible in the video, which does not contradict the physical principle that light arrives before sound. For Wan2.5, the thunder continues for a short duration and gradually attenuates after the lightning has faded, indicating a certain degree of physical plausibility. As for Kling+MMAudio, both thunder events in its generated sample occur after the corresponding lightning appears in the video, likewise not violating the expected physical order. Overall, all three models reflect the light–sound temporal relationship to some extent, though their generated dynamics still show room for improvement when compared with real-world physical behavior.

\subsection{Double channels}

\begin{figure}[t]
    \centering
    \includegraphics[width=1\linewidth]{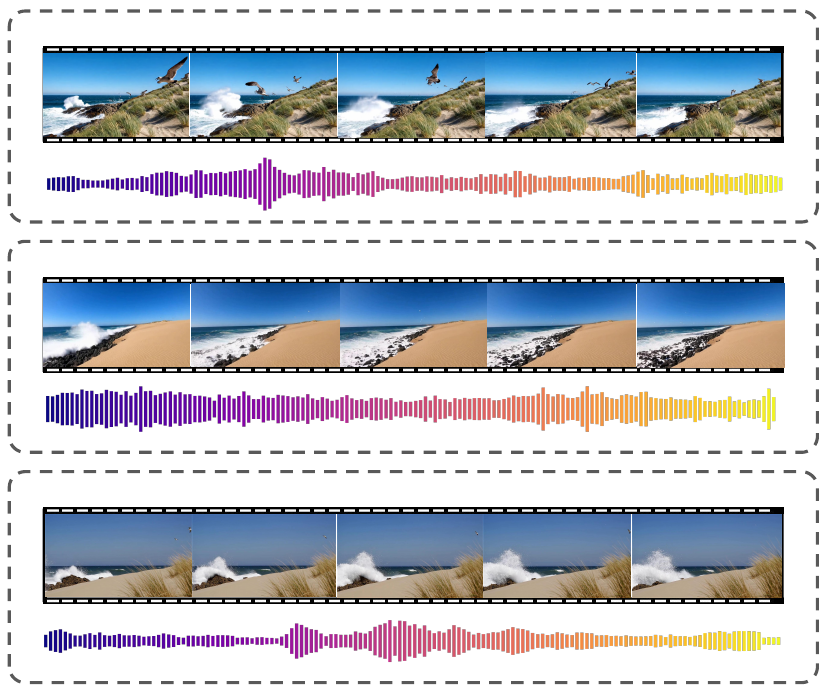}
    \caption{Stereo-sound video for analysis. From top to bottom, the results correspond to the outputs of Veo3, Sora2 and Wan2.5, respectively.}
    \label{fig:dou_show}
\end{figure}

\begin{figure}[t]
    \centering
    \includegraphics[width=1\linewidth]{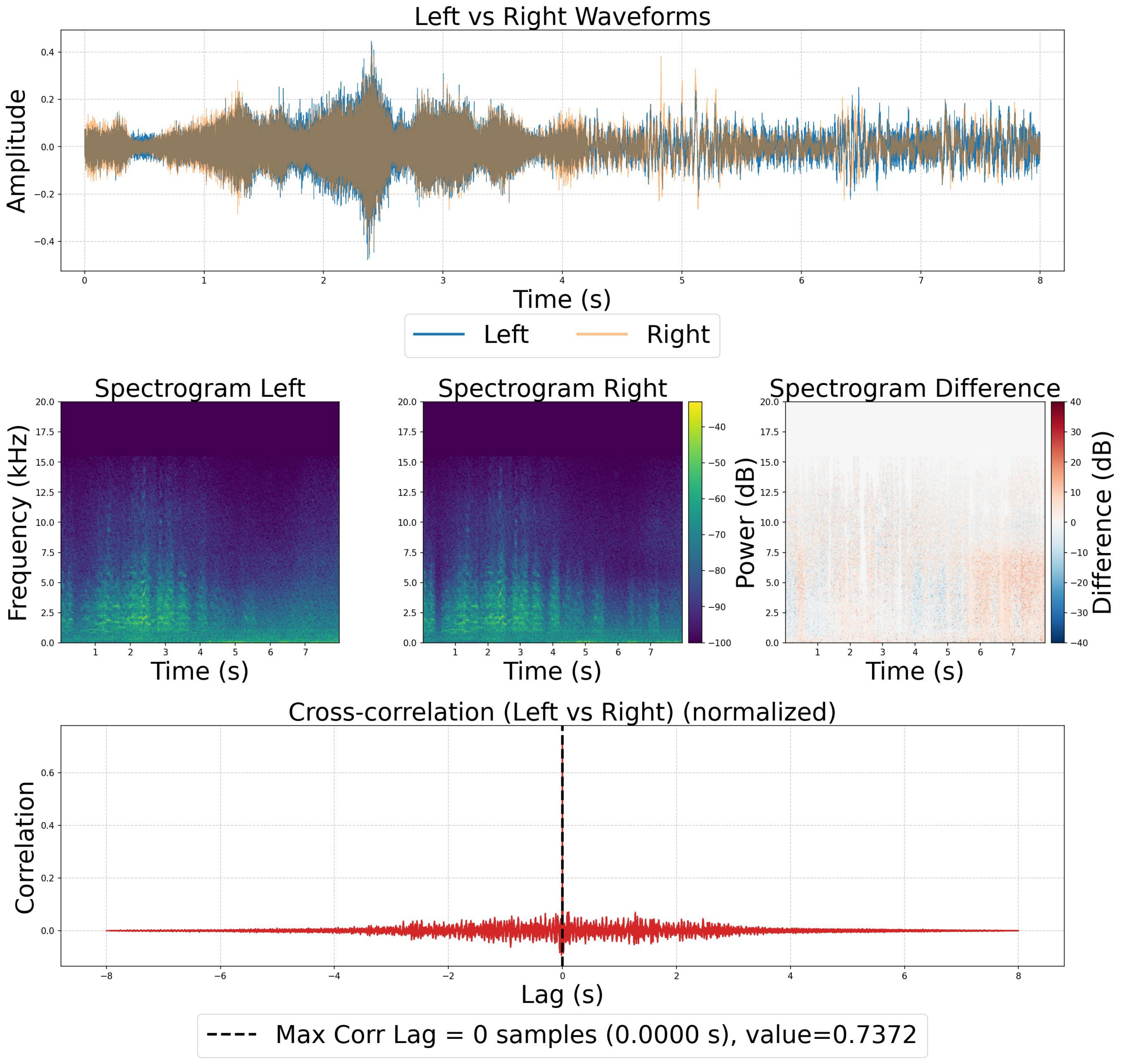}
    \caption{Stereophonic analysis of the video generated by Veo3}
    \label{fig:veo3_stereo}
\end{figure}
\begin{figure}[t]
    \centering
    \includegraphics[width=1\linewidth]{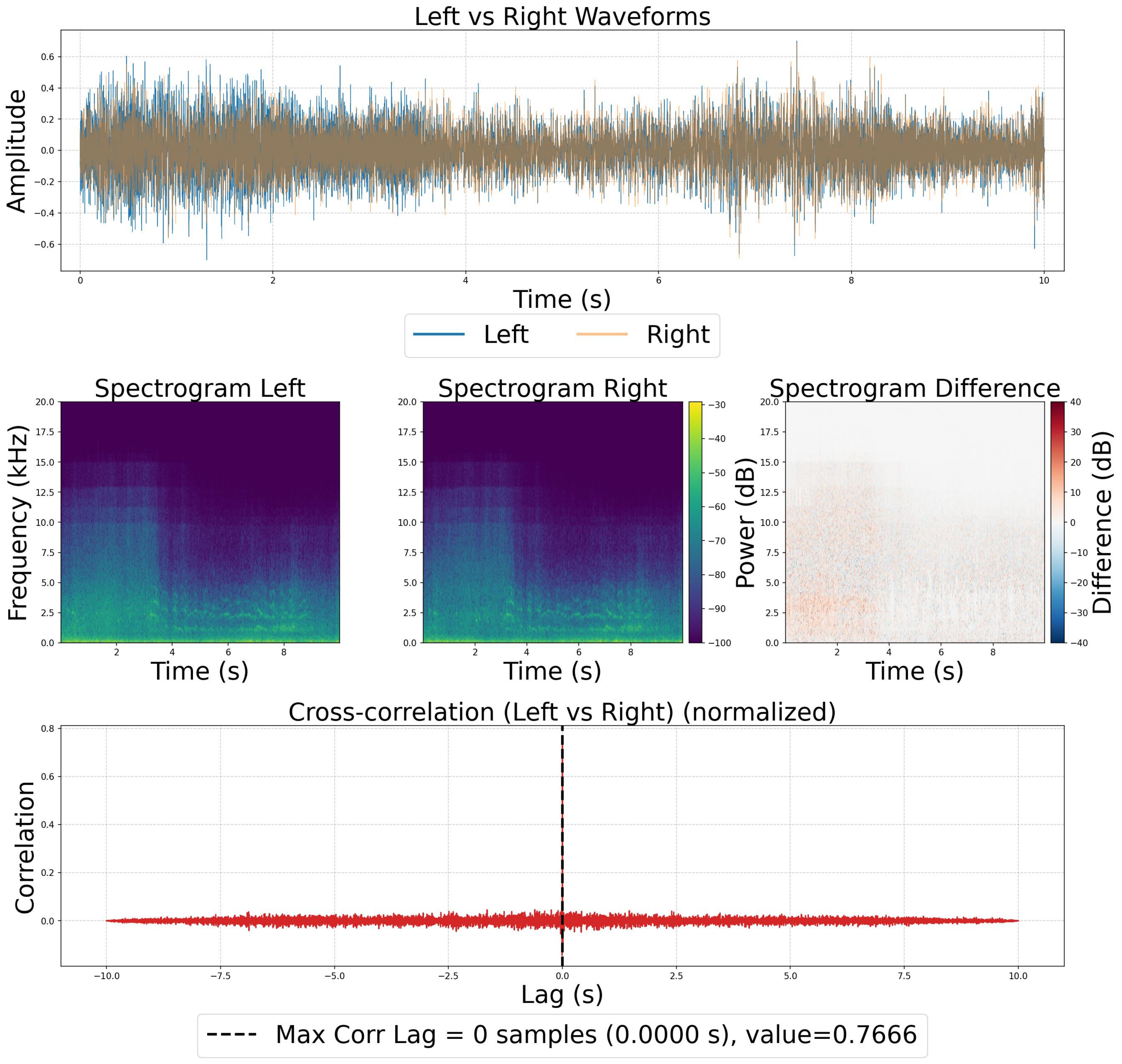}
    \caption{Stereophonic analysis of the video generated by Sora2}
    \label{fig:sora2_stereo}
\end{figure}
\begin{figure}[t]
    \centering
    \includegraphics[width=1\linewidth]{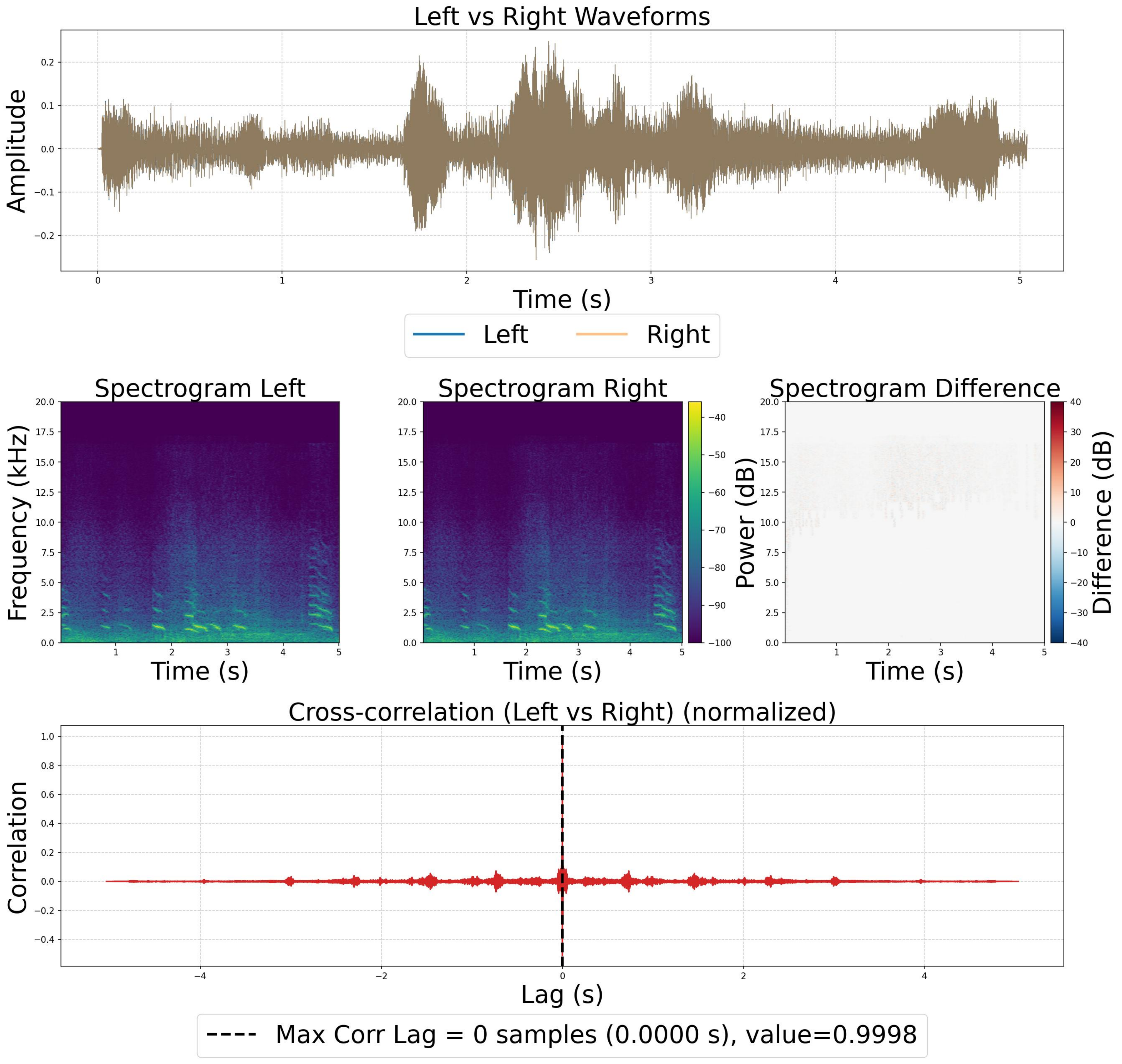}
    \caption{Stereophonic analysis of the video generated by Wan2.5}
    \label{fig:wan2.5_stereo}
\end{figure}

This section evaluates the stereophonic spatial construction capabilities of three AV models. We selected a coastal video example (\cref{fig:dou_show}) and conducted a systematic analysis of the left/right channel waveforms, spectrograms, spectral differences, and cross-correlation characteristics. The prompt used is as follows:

\begin{quote}
    In stereo audio, ocean waves crash powerfully against rocks on the left channel, while seagulls cry and the wind whispers gently through dunes from the right, set under a vast, cloudless blue sky, forming an expansive coastal vista.
\end{quote}

Veo3 (\cref{fig:veo3_stereo}) demonstrates significantly better channel differentiation. The waveform amplitude alternates between channels, and the spectral difference map reveals dynamic energy shifts, confirming the presence of stereophonic information despite high cross-correlation. However, this variation manifests primarily as energy panning rather than the semantically-grounded separation (waves left vs. seagulls right) requested in the prompt. Consequently, while it provides perceptible soundstage movement and depth, the source localization remains ambiguous.

Sora2 (\cref{fig:sora2_stereo}) exhibits nearly identical left and right channels in terms of waveform and frequency spectrum, showing high correlation. This indicates a failure to achieve the specified source separation; ocean waves, seagulls, and wind are blended centrally rather than distributed spatially. Despite perfect synchronization, the output essentially resembles mono audio stored in a dual-channel format, lacking stereophonic width and directionality.

Wan2.5 (\cref{fig:wan2.5_stereo}) shows nearly 100\% channel alignment (correlation value: 0.9998) with consistent spectral characteristics, representing typical mono audio. Consequently, the model did not effectively implement the spatial layout of ``left: waves, right: seagulls,'' resulting in a centralized soundfield lacking perceptible directionality or stereophonic width.

Overall, the performance regarding semantic-driven stereo generation consistent with the prompt indicates substantial room for improvement across all evaluated models. These results highlight that semantic spatial localization remains a significant challenge for current generation frameworks.

\begin{figure}[t]
    \centering
    \includegraphics[width=\linewidth]{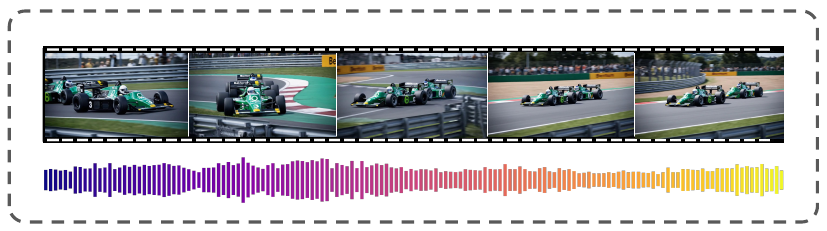}
    \caption{A sample from Veo3’s generated results, illustrating both the Doppler effect and stereophonic audio.}
    \label{fig:veo3_car_show}
\end{figure}

\section{Special samples Analysis}

\begin{figure}[t]
    \centering
    \includegraphics[width=\linewidth]{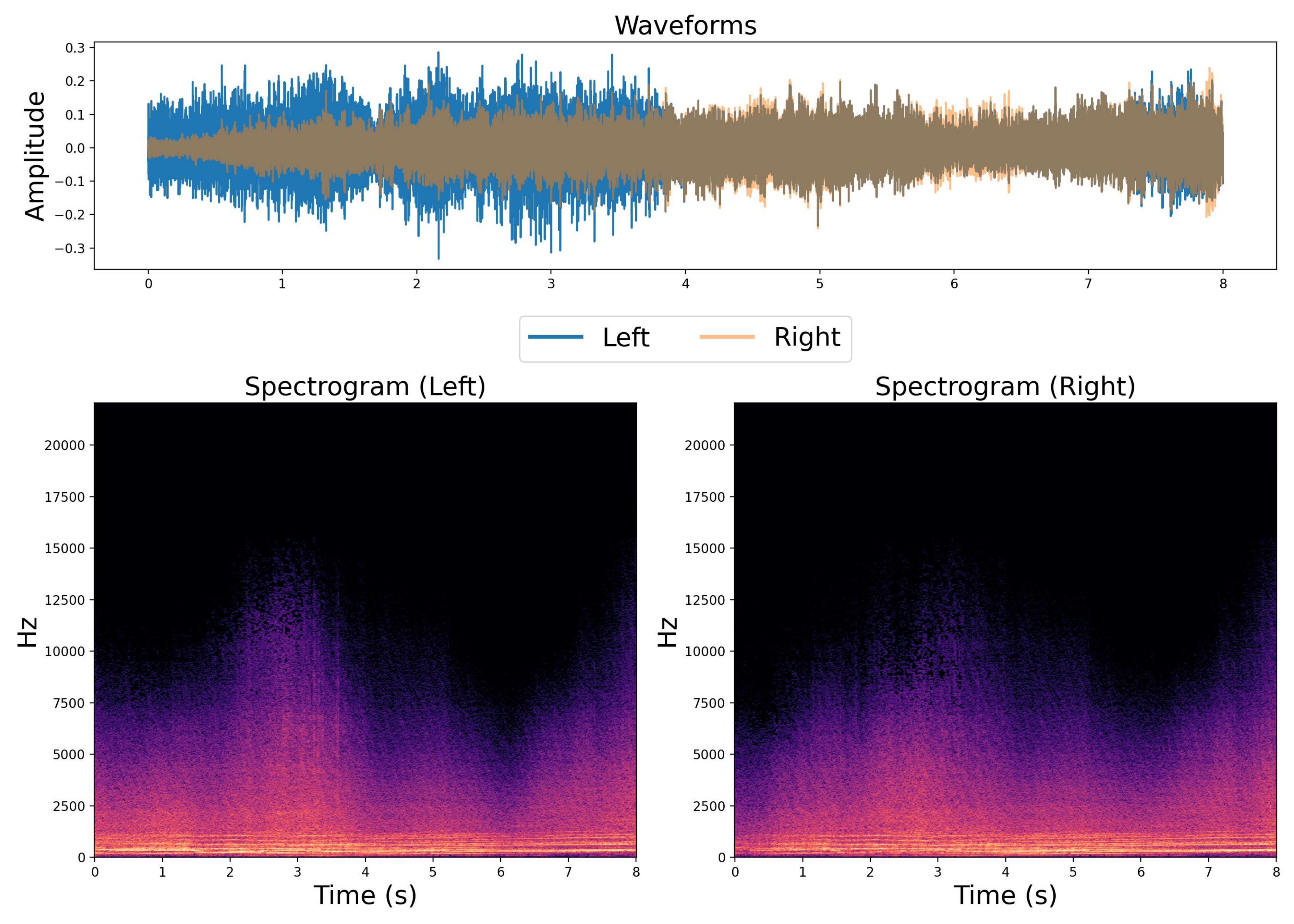}
    \caption{Analysis of the sample generated by Veo3, showing the Doppler effect and left–right channel characteristics.}
    \label{fig:veo3_car}
\end{figure}

\subsection{Veo3 Case Analysis}

We examine a case (\cref{fig:veo3_car_show}) where Veo3 autonomously generated stereophonic audio featuring distinct Doppler effects, notably without explicit spatial specifications in the input prompt. We conducted time-domain waveform and spectrogram analyses for both channels, as shown in \cref{fig:veo3_car}. The specific prompt used is as follows:

\begin{quote}
    On the racetrack, two high-speed racing cars are engaged in an intense competition. The audio should feature the Doppler effect of engine roars changing with the direction and speed of the cars, sharp tire screeches varying rhythmically, and background sounds including crowd cheers and distant wind. The sound field should reflect the distance and positional relationship between the cars, with volume dynamically adjusted as the cars approach or move away from the microphone.
\end{quote}

Spectral analysis confirms the accurate reproduction of physical phenomena. The left-channel spectrogram displays a characteristic Doppler arc—rising from 2s, peaking at 15 kHz near 3.5s, and subsequently descending—accompanied by concentrated high-frequency bursts ($>$10 kHz) during peak intensity (3--4s) that effectively simulate tire friction.

Regarding spatial dynamics, waveform analysis reveals clear left-channel dominance during the 0--4s interval, coinciding with the car's initial visual position. Subjective evaluation further corroborates this synchronization: the auditory frequency modulation and channel balancing align strictly with the vehicle's visual approach and recession, satisfying the prompt's requirements for both physical realism and spatial consistency.

Collectively, this case exemplifies Veo3's capacity to reproduce complex physical phenomena (Doppler effect) and dynamic stereophonic soundfields, highlighting its potential for achieving high physical consistency and spatial accuracy.

\begin{figure}[t]
    \centering
    \includegraphics[width=\linewidth]{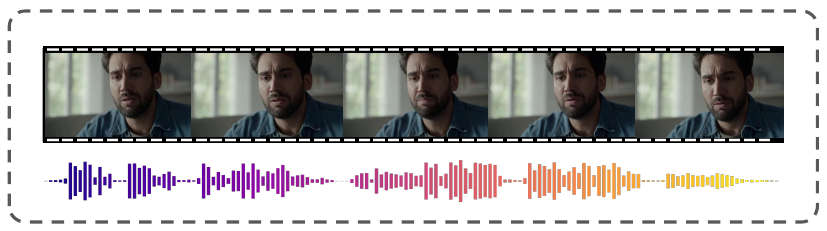}
    \caption{Video generated by Sora2, showing dual-channel audio construction reflecting the intended emotional characteristics.}
    \label{fig:sora2_inner_show}
\end{figure}

\begin{figure}[t]
    \centering
    \includegraphics[width=\linewidth]{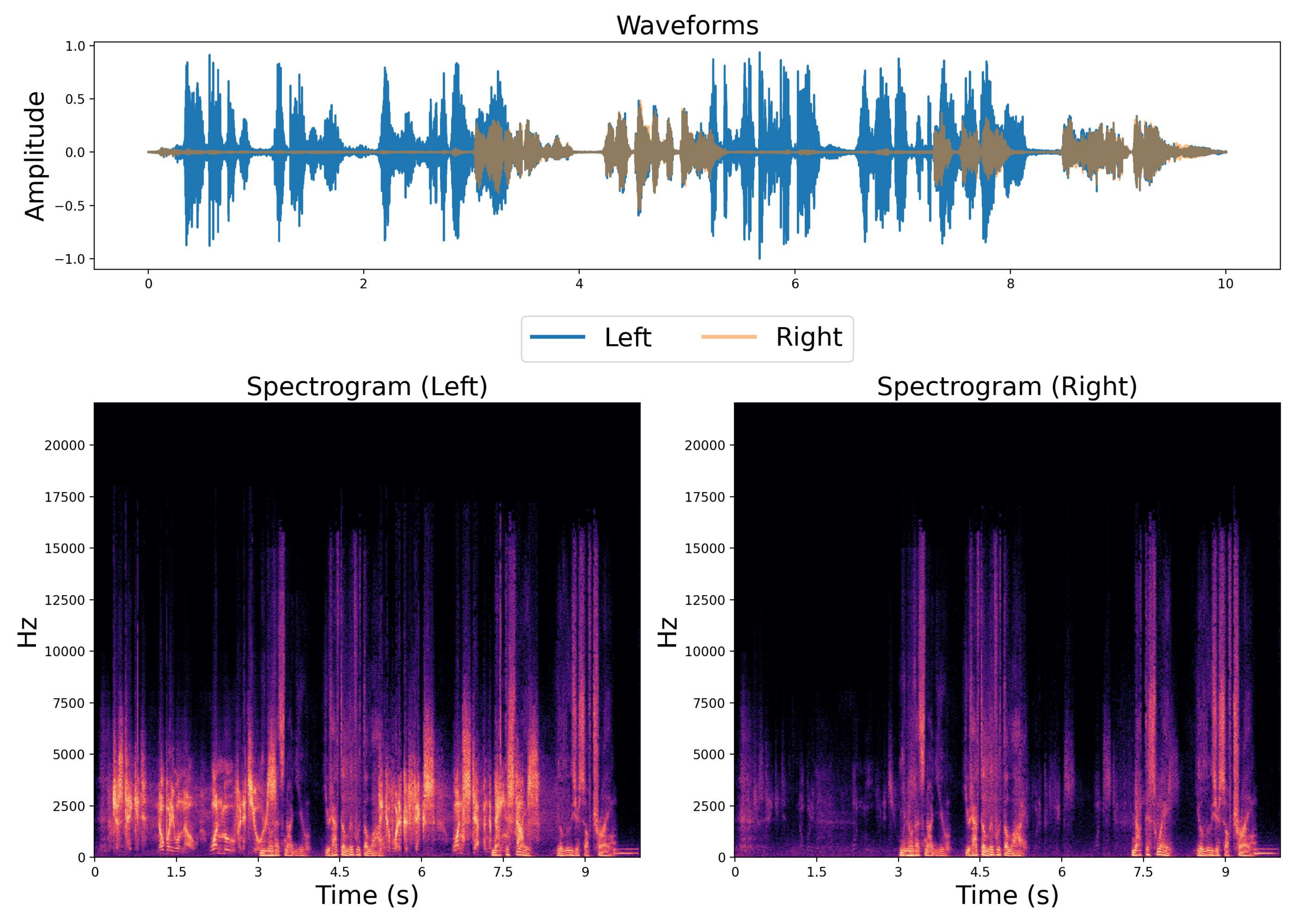}
    \caption{Analysis of the Sora2-generated sample, showing dual-channel emotional rendering and left–right spectral characteristics.}
    \label{fig:sora2_inner}
\end{figure}

\subsection{Sora2 Case Analysis}
This analysis examines a sample generated by Sora2, where the model generated distinct dual-channel audio to capture requested emotional features, despite the absence of explicit spatial constraints. We conducted time-domain waveform and spectrogram analyses for both channels, as illustrated in \cref{fig:sora2_inner}, based on the following prompt:

\begin{quote}
    For a video showing a person's strained facial expressions during a difficult moral decision, create an inner conflict by generating two conflicting layers of whisper-like background audio tracks—one representing temptation and the other conscience—alternating between left and right channels, as if arguing inside the mind.
\end{quote}

The analysis confirms that Sora2 precisely executed the instruction to ``alternate left-right channels.'' Waveform and spectral data reveal clear temporal partitioning: primary emission originates from the left channel (0--3s, 5--9s), alternating with dual-channel activity (3--5s, 9--10s) via staggered, non-overlapping energy bursts. This spatial separation, combined with a rhythmic structure of brief bursts interspersed with silence, effectively simulates the mechanics of a contentious dialogue.

Subjective evaluation further validates that this structure successfully materializes the abstract ``internal debate.'' The model establishes a clear adversarial relationship where a provocative ``temptation'' track and a rational ``conscience'' track alternate in a coherent sequence. The integration of distinct emotional tones with spatial positioning strictly adheres to the prompt's requirements, demonstrating the model's capability to translate psychological conflict into a structured stereophonic narrative.

Collectively, this case exemplifies the model's great potential in integrating technical spatial controllability with coherent emotional narrative.

\begin{figure}[t]
    \centering
    \begin{subfigure}[b]{0.95\linewidth}
        \centering
        \includegraphics[width=\linewidth]{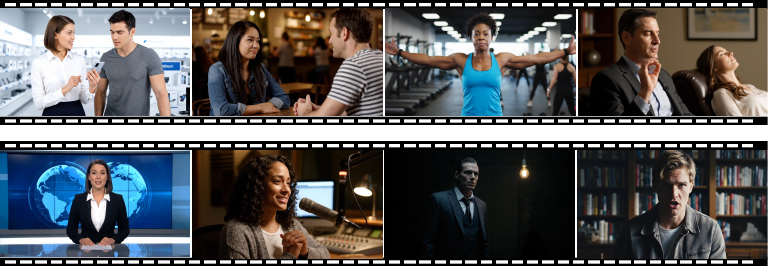}
        \caption{Sample generated by Veo3, showing a strong preference toward Caucasian facial features.}
        \label{fig:dist1}
    \end{subfigure}\\[6pt]
    \begin{subfigure}[b]{0.95\linewidth}
        \centering
        \includegraphics[width=\linewidth]{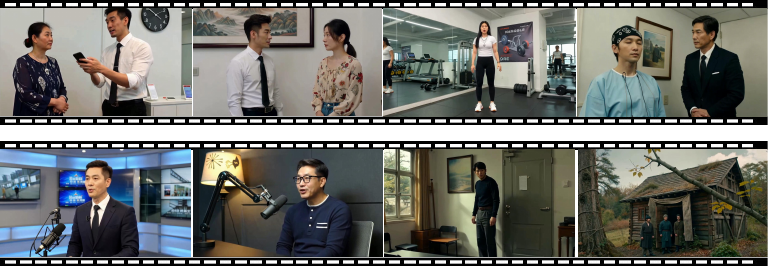}
        \caption{Sample generated by Seedance, demonstrating a tendency to produce subjects with Asian appearances.}
        \label{fig:dist2}
    \end{subfigure}
    \caption{Demographic tendencies in generated human subjects across models. This figure illustrates appearance biases observed during manual inspection.}
    \label{fig:distributions}
\end{figure}

\subsection{Demographic Bias and Data Distribution:}

During our manual inspection of generated samples, we observed a distinct demographic bias in the representation of human subjects across different models. Specifically for example, Veo3 predominantly generates characters with Caucasian features, whereas Seedance exhibits a strong tendency towards generating subjects with Asian appearances. We hypothesize that this disparity is closely correlated with the geographical origins of the models and the implicit distributions of their private training data. Veo3, developed by a US-based entity, likely relies heavily on Western-centric datasets, while Seedance originating from Asian developers, likely incorporate a higher proportion of Asian-centric data. This observation suggests that generative models tend to reflect the demographic characteristics inherent in their training corpora.Video screenshots illustrating this bias can be seen in \cref{fig:distributions}.

%-------------------流程画图展示+
\section{MLLM Based Evaluation Cases}
\subsection{Macro Evaluation System Prompt Sample}
As introduced in the main paper, our evaluation framework leverages Qwen2.5 Omni 7B~\cite{xu2025qwen2} to provide a scalable and standardized alternative to traditional MOS. This supplementary section provides the specific implementation details for the coarse-grained (macro) evaluation level. To achieve this, we design a suite of detailed system prompts. Each prompt casts the MLLM into the role of a specialized expert and provides a comprehensive, five-point scoring rubric (scored 1-5) and output requirements. This methodology ensures that the MLLM's assessment is constrained, consistent, and targeted to a specific quality dimension. Below, we provide two examples from our macro-evaluation prompt suite.

\vspace{10pt}
\noindent \textbf{Macro - Visual-Realism:}

\begin{quote}

You are a Visual Realism Analyst. Assess whether the video obeys real-world physics, material behavior, and human visual perception.

    Evaluate these five core aspects:
    1. Object permanence \& occlusion: Objects should not appear/disappear abruptly; when one object passes behind another, it must be partially hidden consistently.
    2. Biomechanically plausible motion: Human/animal movement must respect joint limits, weight, and momentum (e.g., no floating limbs, unnatural gait, or instant direction changes).
    3. Physically consistent rendering: Lighting, shadows, color temperature, and perspective must align with a single, coherent light source and spatial layout (e.g., shadows should point away from light, parallel lines converge correctly).
    4. Temporal coherence: Motion must be smooth across frames—no sudden jumps, speed glitches, or inconsistent frame-to-frame transitions without physical cause.
    5. Material \& environmental interaction: Objects should respond realistically to forces and surroundings (e.g., fabric drapes, water splashes on impact, footsteps deform soft ground, or glass reflects surroundings).

    Use this scoring scale:
    5: Perfect realism — every frame respects physics, perception, and material behavior. No anomalies detected.
    4: Minor, brief flaws — e.g., a shadow slightly misaligned for one frame, or a limb briefly stiff. Does not break believability.
    3: Noticeable but isolated issues — e.g., a character walks with robotic knees, an object briefly “pops” into place, or water fails to splash on impact. Realism is weakened but still functional.
    2: Frequent violations — e.g., objects teleport, shadows flip direction, joints bend impossibly, or motion stutters unnaturally. Disrupts immersion consistently.
    1: Physically incoherent — chaotic visuals: people vanish mid-step, lighting shifts randomly, perspective collapses, or materials behave like abstract textures. Feels like broken CGI or hallucination.

    Output Requirements:
    - Return ONLY a single JSON object.
    - Must contain exactly two keys: ``score" (integer 1-5) and ``reason" (string, $\ge$15 characters).
    - In ``reason", cite at least one specific anomaly with approximate timestamp (e.g., ``t 0:12, the chair reappears after being fully occluded by a person") and explain how it violates realism.
    - Do NOT include markdown, extra text, or additional fields.

    Example valid output:
    {``score": 2, ``reason": ``At 0:09, the character's elbow bends backward during a reach, violating joint biomechanics."}

\end{quote}

\noindent \textbf{Macro - Expressiveness:}

\begin{quote}

    You are a Narrative Analyst. Evaluate how effectively the audio supports the video's emotional tone and storytelling.

    Focus on two key dimensions:
    1. Emotional alignment: Does the sound (music, effects, silence, etc.) match the intended mood—such as tension, joy, grief, or suspense—at each moment?
    2. Narrative function: Does audio actively clarify or enhance the story? Examples include:
    - Highlighting a key action (e.g., a heartbeat during a reveal)
    - Conveying character perspective (e.g., muffled sound during dazed POV)
    - Bridging scenes through sound continuity (e.g., train whistle fading into next location)
    - Providing off-screen context (e.g., distant sirens implying danger)

    Use this scoring scale:
    5: Exceptional narrative and emotional synergy — audio is integral to the story, powerfully shaping mood and meaning (e.g., silence used as dramatic punctuation, sound design reveals inner state).
    4: Strong support — clear emotional match and helpful narrative cues; enhances understanding and immersion without being revolutionary.
    3: Minimal contribution — audio is present but generic or neutral (e.g., ambient pad with no emotional inflection); neither helps nor hurts significantly.
    2: Misaligned or confusing — emotional tone clashes with visuals (e.g., upbeat music over a funeral) or omits critical cues (e.g., silence during a pivotal line).
    1: Actively harmful — audio contradicts the scene's intent or creates narrative chaos (e.g., laugh track over violence), impairing viewer comprehension.

    Output Requirements:
    - Return ONLY a single JSON object.
    - Must contain exactly two keys: ``score" (integer 1-5) and ``reason" (string, $\ge$15 characters).
    - In ``reason", cite at least one specific moment with approximate timestamp (e.g., ``At 0:14, tense strings swell as the character reaches for the gun, amplifying suspense") and explain how it supports or undermines story/emotion.
    - Do NOT include markdown, extra text, or additional fields.

    Example valid output:
    {``score": 4, ``reason": ``At 0:22, muffled audio during the car crash effectively conveys the protagonist's disorientation and enhances emotional impact."}

\end{quote}

\subsection{Micro Evaluation Sample Display}
This section demonstrates the concrete implementation of the QA scoring workflow under the micro-level metric framework, using one video example from each of the T2AV and I2AV tasks, with outputs generated by one AV model and one V+A model for evaluation.

\noindent \textbf{Example for T2AV:}

The sampled prompt is: In the quiet streets of a residential neighborhood at dawn, the distant rumble of the first bus engine echoes in the background. Nearby, an early-rising bird chirps. The milkman places glass bottles into a doorstep crate, producing a soft clinking sound. A jogger runs past in athletic shoes, their footsteps making a rhythmic slap-slap noise. 

\vspace{1em}
\noindent \textit{\textbf{Veo3 (Overall Score 0.8):}}

\begin{figure}[ht]
    \centering
    \includegraphics[width=\linewidth]{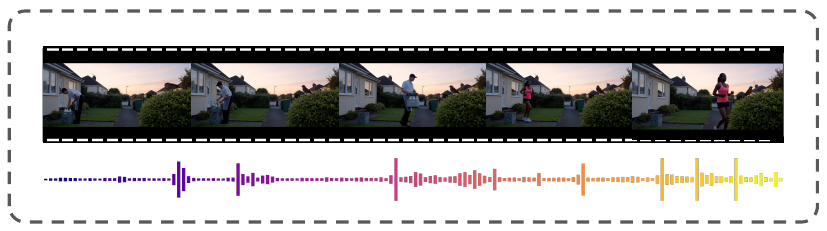}
    \caption{Video example generated by Veo3 on the T2AV task.}
    \label{fig:veo3_qa_samp}
\end{figure}

\texttt{Q:Is the distant rumble of the bus engine audible in the background, creating an echo effect?}

\texttt{A:No: The distant rumble of the bus engine is not audible in the background.}

\texttt{Q:Can the chirping of the early-rising bird be heard clearly amidst the quiet streets?}

\texttt{A:Yes: The chirping of the early-rising bird is clearly audible.}

\texttt{Q:Does the sound of glass bottles clinking softly occur when the milkman places them into the doorstep crate?}

\texttt{A:Yes: The sound of glass bottles clinking softly occurs when the milkman places them into the doorstep crate.}

\texttt{Q:Are the rhythmic slap-slap noises of the jogger's footsteps audible as they run past?}

\texttt{A:Yes: The rhythmic slap-slap noises of the jogger's footsteps are audible as they run past.}

\texttt{Q:Is the overall soundscape predominantly quiet, with only the described sounds being prominent?}

\texttt{A:Yes: The audio captures the described sounds of a quiet residential neighborhood at dawn, including the distant rumble of a bus engine, a bird chirping, the milkman's clinking bottles, and the jogger's footsteps.}

\vspace{1em}
\noindent \textit{\textbf{Seedance+MMAudio (Overall Score 0.4):}}

\begin{figure}[ht]
    \centering
    \includegraphics[width=\linewidth]{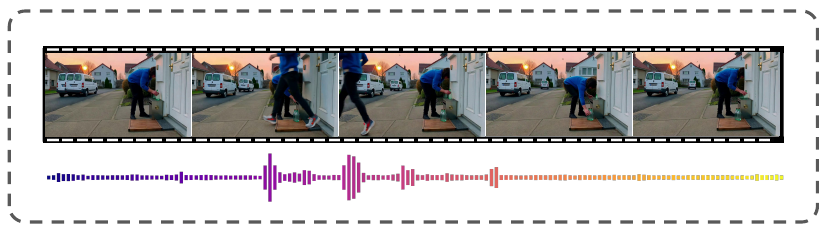}
    \caption{Video example generated by Seedance+MMAudio on the T2AV task.}
    \label{fig:seedmm_qa_samp}
\end{figure}

\texttt{Q:Is the distant rumble of the bus engine audible in the background, creating an echo effect?}

\texttt{A:No: The distant rumble of the bus engine is not audible in the background, nor is there an echo effect.}

\texttt{Q:Can the chirping of the early-rising bird be heard clearly amidst the quiet streets?}

\texttt{A:No: The chirping of the early-rising bird is not clearly audible in the audio.}

\texttt{Q:Does the sound of glass bottles clinking softly occur when the milkman places them into the doorstep crate?}

\texttt{A:Yes: The sound of glass bottles clinking softly occurs when the milkman places them into the doorstep crate.}

\texttt{Q:Are the rhythmic slap-slap noises of the jogger's footsteps audible as they run past?}

\texttt{A:Yes: The rhythmic slap-slap noises of the jogger's footsteps are audible as they run past.}

\texttt{Q:Is the overall soundscape predominantly quiet, with only the described sounds being prominent?}

\texttt{A:No: The soundscape includes traffic noise and footsteps, which are not described in the reference.}

\noindent \textbf{Example for I2AV:}

The sampled prompt is: On a large outdoor music festival night, lights flicker on stage as the music pulses powerfully. In the foreground, the crowd's cheers intertwine with the background music, creating a rich sound field. The rhythm is intense, with deep, resonant bass that spreads from the center of the stage throughout the venue, accompanied by lighting changes. Environmental sounds include the footsteps of the crowd, occasional cheers, and applause, building up an enthusiastic atmosphere. The timbres are diverse, with high-frequency electronic tones contrasting sharply against low-frequency rhythms, forming a clearly layered sound field where foreground and background sound sources are distinctly separated.

\vspace{1em}
\noindent \textit{\textbf{Wan2.5 (Overall Score 0.8):}}

\begin{figure}[ht]
    \centering
    \includegraphics[width=\linewidth]{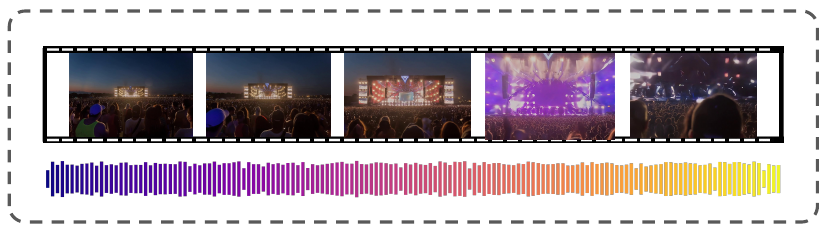}
    \caption{Video example generated by Wan2.5 on the I2AV task.}
    \label{fig:wan25_qa_samp}
\end{figure}

\texttt{Q:Does the audio include powerful music with deep, resonant bass that spreads from the center of the stage throughout the venue?}

\texttt{A:Yes: The audio includes powerful music with deep, resonant bass that spreads from the center of the stage throughout the venue.}

\texttt{Q:Are the crowd's cheers and applause audible and intertwined with the background music, contributing to a rich sound field?}

\texttt{A:Yes: The crowd's cheers and applause are audible and intertwined with the background music, contributing to a rich sound field.}

\texttt{Q:Is there a clear distinction between high-frequency electronic tones and low-frequency rhythms, forming a layered sound field?}

\texttt{A:Yes: The audio clearly distinguishes between high-frequency electronic tones and low-frequency rhythms, forming a layered sound field.}

\texttt{Q:Is Do the environmental sounds include footsteps of the crowd that contribute to the enthusiastic atmosphere?}

\texttt{A: Yes: The video's audio includes the footsteps of the crowd, contributing to the enthusiastic atmosphere.}

\texttt{Q:Are the lighting changes synchronized with the rhythm of the music?}

\texttt{A:No: The lighting changes are not synchronized with the rhythm of the music.}

\vspace{1em}
\noindent \textit{\textbf{Kling+ThinkSound (Overall Score 1.0):}}
\begin{figure}[ht]
    \centering
    \includegraphics[width=\linewidth]{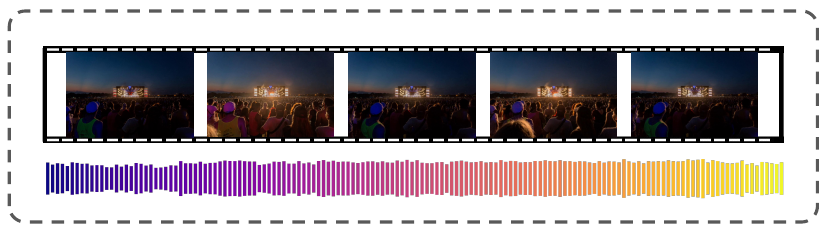}
    \caption{Video example generated by Kling+ThinkSound on the I2AV task.}
    \label{fig:kth_qa_samp}
\end{figure}

\texttt{Q:Does the audio include powerful music with deep, resonant bass that spreads from the center of the stage throughout the venue?}

\texttt{A:Yes: The audio includes powerful music with deep, resonant bass that spreads from the center of the stage throughout the venue.}

\texttt{Q:Are the crowd's cheers and applause audible and intertwined with the background music, contributing to a rich sound field?}

\texttt{A:Yes: The crowd's cheers and applause are audible and intertwined with the background music, contributing to a rich sound field.}

\texttt{Q:Is there a clear distinction between high-frequency electronic tones and low-frequency rhythms, forming a layered sound field?}

\texttt{A:Yes: The audio clearly distinguishes between high-frequency electronic tones and low-frequency rhythms, creating a layered sound field.}

\texttt{Q:Is Do the environmental sounds include footsteps of the crowd that contribute to the enthusiastic atmosphere?}

\texttt{A:Yes: The video includes the footsteps of the crowd, contributing to the enthusiastic atmosphere.}

\texttt{Q:Are the lighting changes synchronized with the rhythm of the music?}

\texttt{A:Yes: The lighting changes are synchronized with the rhythm of the music.}

\end{document}